\documentclass{sig-alternate}
\usepackage[ruled]{algorithm2e}
\usepackage{subfigure}
\usepackage{multirow}
\usepackage{comment}
\newtheorem{definition}{Definition}
\begin{document}
%
\conferenceinfo{xxx}{xxxxx}

\title{Less is More: Learning Prominent and Diverse Topics for Data Summarization}
%
%
%
%

\author{
	Jian Tang$^1$, Cheng Li$^2$, Ming Zhang$^{1}$, Qiaozhu Mei$^2$ \\
	\affaddr{$^1$School of EECS, Peking University, \{tangjian, mzhang\}@net.pku.edu.cn }\\
		\affaddr{$^2$School of Information, University of Michigan,  \{lichengz, qmei\}@umich.edu}\\
}

\maketitle
\begin{abstract}
Statistical topic models efficiently facilitate the exploration of large-scale data sets. Many models have been developed and broadly used to summarize the semantic structure in news, science, social media, and digital humanities. However, a common and practical objective in data exploration tasks is not to enumerate all existing topics, but to quickly extract representative ones that broadly cover the content of the corpus, i.e., a few topics that serve as a good summary of the data. Most existing topic models fit exactly the same number of topics as a user specifies, which have imposed an unnecessary burden to the users who have limited prior knowledge. We instead propose new models that are able to learn fewer but more representative topics for the purpose of data summarization. We propose a reinforced random walk that allows prominent topics to absorb tokens from similar and smaller topics, thus enhances the diversity among the top topics extracted. With this reinforced random walk as a general process embedded in classical topic models, we obtain \textit{diverse topic models} that are able to extract the most prominent and diverse topics from data. The inference procedures of these diverse topic models remain as simple and efficient as the classical models. Experimental results demonstrate that the diverse topic models not only discover topics that better summarize the data, but also require minimal prior knowledge of the users. 

\end{abstract}


\section{Introduction}
A huge amount of unstructured data has been continuously generated from various online information sources at an unprecedented speed, usually referred to as the ``big data.'' A major challenge for data scientists has emerged along with this trend, which is concerned with how to facilitate the understanding and exploration of the big data. 

Statistical topic models \cite{blei2012probabilistic}, e.g., the probabilistic latent semantic analysis (PLSA)~\cite{hofmann1999probabilistic} and the latent Dirichlet allocation (LDA)~\cite{blei2003latent}, have been widely recognized as effective tools to assist the real users in understanding and exploring the data. This family of probabilistic models are designed to automatically infer the hidden themes (a.k.a.\ topics) that are salient in the data collection. 
The discovered topics can be further utilized in other data mining tasks such as classification~\cite{lacoste2008disclda,zhu2009medlda}, clustering~\cite{karandikar2010clustering, ramage2009clustering}, sentiment analysis~\cite{mei2007topic,lin2009joint}, and user modeling~\cite{ahmed2011scalable}.

These models commonly rely on a strong assumption that the user knows the actual number of topics in the data. This assumption may hold for small and restricted data sets, but becomes impractical when the data is big and when the domain is open, thus has limited the usefulness of topic models in practical data exploration tasks. Sophisticated treatments have to be applied to topic models in order to relax this assumption, which lead to various nonparametric \cite{teh2006hierarchical} and hierarchical versions of the models \cite{li2006pachinko} that are much more complicated. While these treatments successfully advanced the theoretical understanding of topic models, they have not completely solved the problem. In practice, the quality of topics extracted by nonparametric models is usually compromised, while how to find the hierarchical structure of topics remains mysterious.

Clearly, all the aforementioned challenges are due to the practice that we wanted to extract \textbf{all} the topics in the data. Is this, however, a necessary practice at all? We find that in many real world scenarios, the user actually wants to explore a few \textit{most prominent topics} out there, instead of enumerating \textit{all} possible topics in the data. Indeed, when searching for news reports about an event, although an investigator may still want to enumerate every aspect or detail of the event, an ordinary Web user only needs to read a few articles that well summarize the major aspects of the event. When exploring the literature of a new field, a researcher may start with digesting several major trends of research instead of investigating every research topic in that field. That says, different from investigation tasks where the concern is to enumerate \textbf{all} the topics in a data set, in most data exploration tasks the concern is to find the \textbf{top K} topics that reasonably \textit{summarize}, or cover the entire data collection. This is analogical to text summarization tasks where a summary of a limited length is created to cover the most important points of the original document(s), or to information retrieval tasks where a limited number of results are presented to represent the big picture of content relevant to an information need. Comparing to the conventional applications of topic models, the number K in this practice does not rely on the impractical assumption about the user's prior knowledge, but rather depends on the budget or the personal need of the user (e.g., I have time to digest three research topics, or I want to know the two most representative perspectives in a debate). In practice, this number, K, can be much smaller than the actual (unknown) number of topics in the data. 

As a good summary of the data collection, the K topics extracted should not only be meaningful and interpretable, but also cover as much content of the original collection as possible. This naturally requires the topics extracted to be the most prominent ones in the data collection, while at the meantime they should cover diverse aspects of the collection. A straightforward way to generate such a summary is to simply fit exactly K topics through a classical topic model. However, fitting only a few topics to a big data collection is likely to under-fit the data, making the extracted topics less meaningful or interpretable.  


An alternative approach may first fit a large number of topics to the data and then pick the most important ones, e.g., the ones with the largest sizes. These topics are likely to be meaningful to the user. However, learning too many topics runs the risk of over-fitting the data, making the extracted topics either too small or too similar to each other. As a result, even the top ones may only cover partial aspects of the collection. This challenge motivates us to investigate new ways to extract the most representative, meaningful, and yet non-redundant topics from the data. 



\begin{figure}[htdb!]
\centering
\subfigure[\small A network of topics. Nodes are topics extracted by a classical topic model; topics with a high similarity are connected through an edge.]{
\includegraphics[width=1.5in]{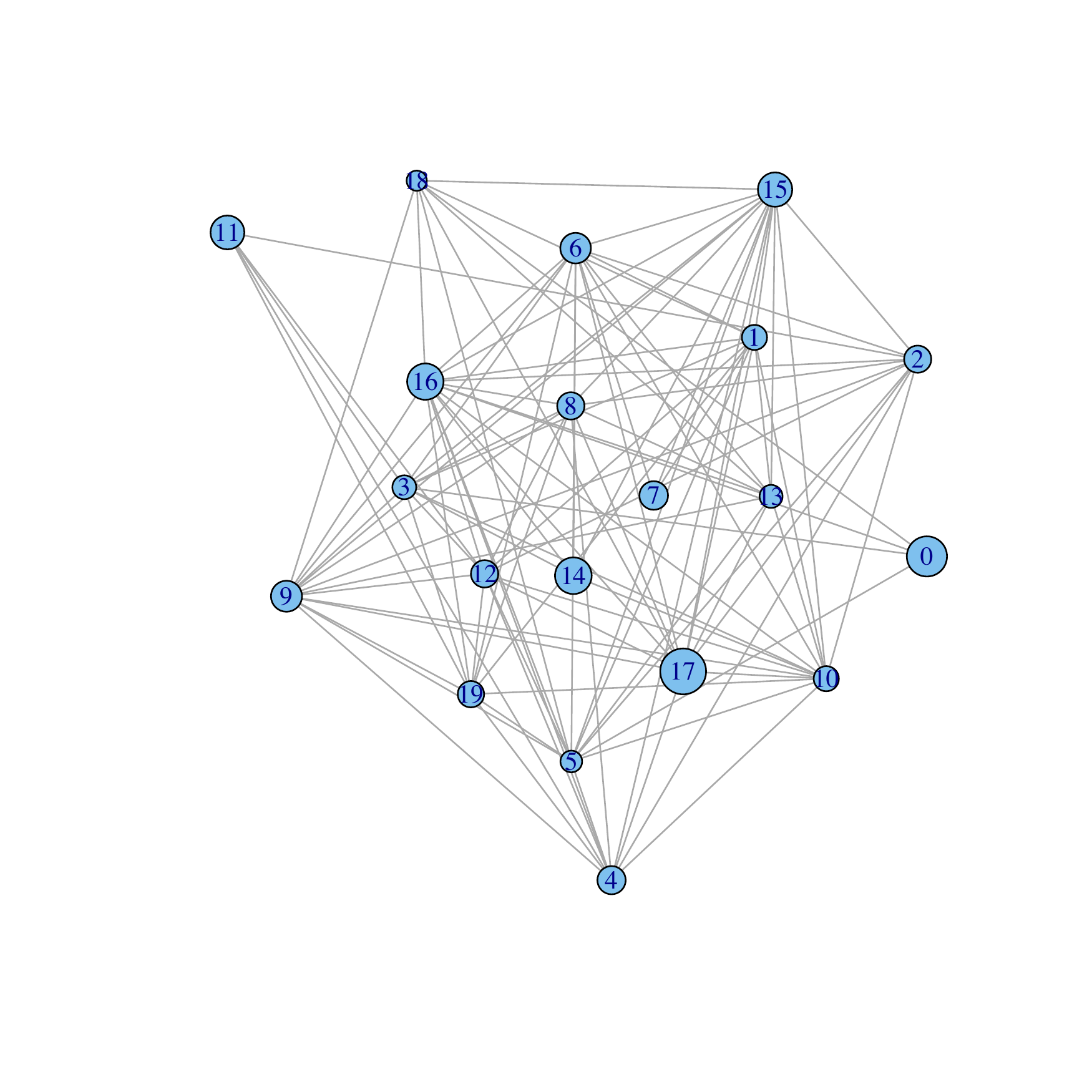}
\label{fig:network-plsa}
}
\subfigure[\small Network of topics after the reinforced random walk. A few prominent and non-redundant topics stand out and present a good coverage of the network.]{
\includegraphics[width=1.5in]{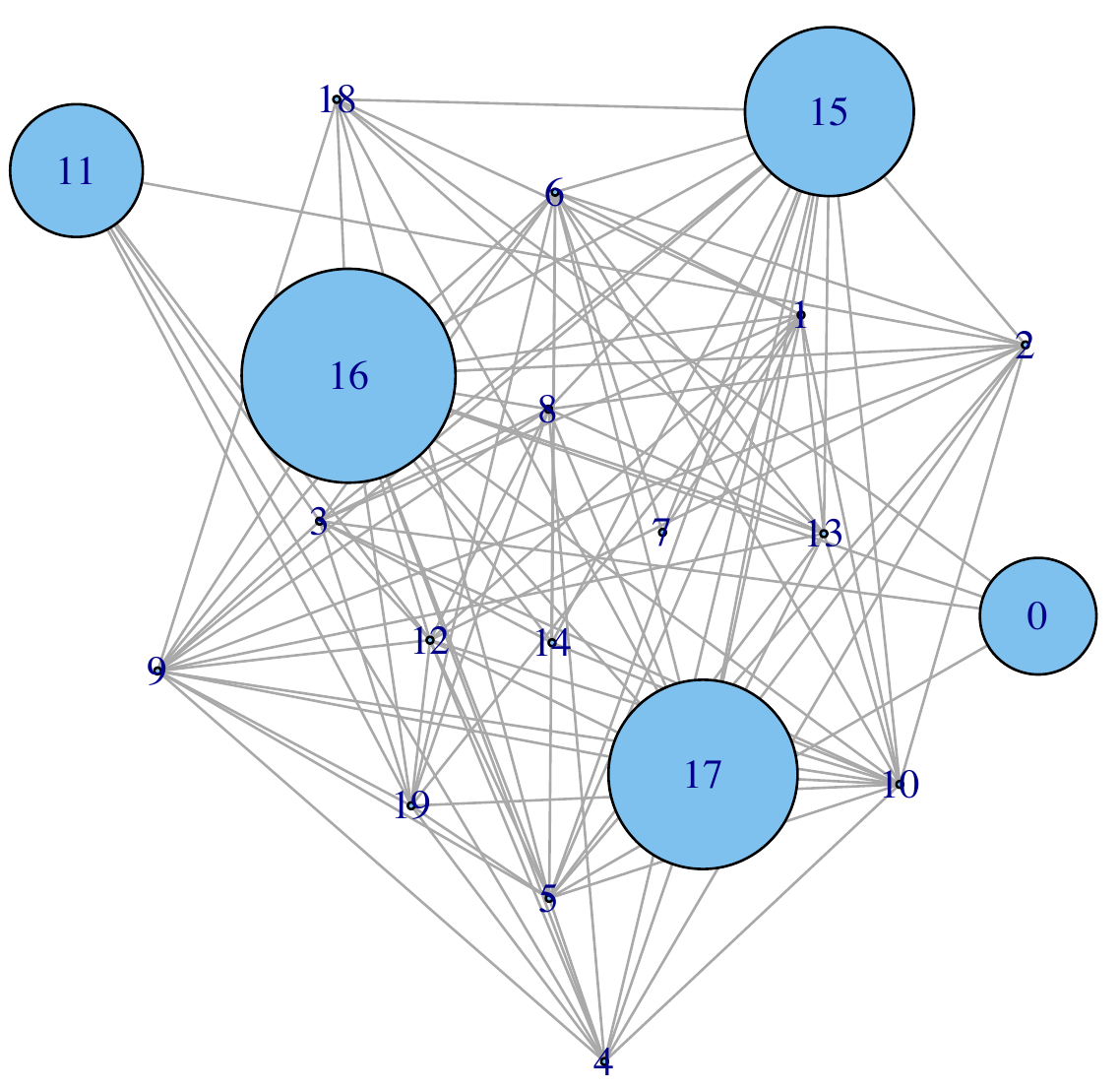}
\label{fig:network-divplsa}
}
\caption{Reinforced Random Walk on a ``Social'' Network of Topics.}
\label{fig:intro-topics-comparsion}
\end{figure}

In this paper, we propose a \emph{reinforced random walk} among a ``social network'' of the topics that allows prominent topics to \textit{absorb} tokens from similar and smaller topics.  
Specifically, during the inference procedure, a topic network is built to model the correlations, or interconnections among the topics. For each word token in the collection, after it is assigned to one of the topics, it is allowed to take a random walk in the topic network, and possibly transit to other topics. The probability of transition from one topic to another is initialized based on the similarity between the topics and then \emph{reinforced} by the size of the target topic, i.e., the number of tokens already assigned or transited to it. During this process, the tokens belonging to the smaller topics are likely to transit to the larger and similar topics. In other words, the larger topics will \emph{absorb} resource from its local neighborhood (of similar but smaller topics). While the process continues, a few prominent topics stand out from their local neighborhood and become the most representative ones in the topic network. At the meantime, these topics tend to represent different neighborhoods in the network, thus enhances the diversity among the top topics (see Figure~\ref{fig:intro-topics-comparsion}). 

We embedded this \emph{reinforced random walk} process into the classical topic models PLSA and LDA, and obtained two \textit{diverse topic models} DivPLSA (abbreviation for diverse PLSA) and DivLDA(abbreviation for diverse LDA). The learning procedures of the two models remain as simple and efficient as the classical ones but are able to learn the prominent and diverse topics pervading the data. Top topics extracted by the diverse topic models are not only meaningful but also serving as good summaries of the collection. Experimental results using four real-world datasets demonstrate the effectiveness of the two diverse topic models for data summarization, compared with classical topic models and alternative approaches.

\noindent{\textbf{Organization.}} The rest of this paper is organized as follows. Section 2 discusses the related work. Section 3 formally defines the problem of topic modeling for data summarization. Section 4 introduces the reinforced random walk and the proposed diverse topic models. The effectiveness of the diverse topic models are empirically evaluated in Section 5, and the study is concluded in Section 6.

\section{Related Work}
Our work presents a good analogy to extractive text summarization \cite{erkan2004lexrank}, the goal of which is to construct a short summary of an individual document or multiple documents by extracting the most representative sentences in the document(s). Instead of summarizing a single document or a few documents with sentences, topic modeling is used to summarize a large document collection with latent topics pervading the collection, in which the topics serve as the role of ``sentences'' in document summarization. Analogical to text summarization, the coverage of the extracted topics in the original data should be maximized and the redundancy among the extracted topics should be minimized. In other words, the extracted topics should be both prominent and diverse. The diversity among the results has been recognized to be important in not only text summarization~\cite{carbonell1998use}, but also many other applications including Web search engines~\cite{agrawal2009diversifying}, query suggestion~\cite{ma2010diversifying}, ranking in information networks~\cite{mei2010divrank}, and recommender systems~\cite{ziegler2005improving}.   

Many algorithms for enhancing the diversity of the results of a retrieval or a mining system have been proposed. In~\cite{carbonell1998use}, Carbonell et al.\ proposed the Maximal Marginal Relevance (MMR) criterion for the problem of document retrieval, which aims to reduce the redundancy among the retrieved documents while preserving their relevance to the query. The MMR algorithm greedily selects the document that has the largest marginal relevance to the query, which is the relevance of the document to the query penalized by the largest similarity between the document and the selected documents. In~\cite{zhu2007improving}, a ``soft'' version of MMR algorithm called Grasshopper based on absorbing random walk is proposed. In~\cite{mei2010divrank}, Mei et al.\ proposed a unified process DivRank to generate the entire ranked list of vertices in information networks by balancing the prestige and diversity. The DivRank algorithm is built on top of the vertex-reinforced random walk, in which the transition probabilities between the vertices are reinforced by the number of visits to the target vertex.  Neighboring vertices are competing resources against each other, and finally the probability mass is distributed among some diverse vertices. Our proposed \emph{reinforced random walk} shares similar idea with the DivRank algorithm. Instead of reinforced by the number of visits to each vertex (topic in our case), the transition probabilities of the random walk by the word tokens are reinforced by the size of the topics. By doing this the larger topics would gradually absorb tokens from the smaller similar ones and end up with the most prominent and diverse topics in the data.

Another related direction is to learn diversified mixture components in area of mixture models ~\cite{DBLP:conf/nips/PetraliaRD12, zou2012priors}. In most of the existing mixture models, the mixture components are identically drawn from a prior distribution. When the data is over-fitted, multiple mixture components are used to model one underlying group in the data, which results in many redundant mixture components. In~\cite{DBLP:conf/nips/PetraliaRD12}, Petralia and Rao proposed a repulsive mixture prior which penalizes redundant components. Zou and Adams replaced the underlying i.i.d.\ prior for mixture components with a determinantal point process (DPP), which defines a probability measure over the entire set of probability distributions~\cite{zou2012priors}. However, though these models are able to obtain more diverse mixture components than classical mixture models, multiple similar mixture components are still used to model one group in the data.  In our proposed diverse topic models, the larger mixture components are able to fully absorb the smaller duplicated ones, and this indicates our models can reduce the number of unnecessary mixture components. 

\section{Problem Definition}
Statistical topical models represent and summarize the data through the discovery of hidden topics pervading the data collection. Each topic is formally defined as follows:

\begin{definition}
\label{def:topic}
\textsl{A \textbf{topic} $\phi$ is defined as a multinomial distribution over words in a vocabulary $V$,
i.e. $\{p(w|\phi)\}_{w\in V}$. Without loss of generality, we assume there are $\mathcal{K}$ topics in total in a given data collection, where $\mathcal{K}$ is unknown.  
}
\end{definition}

In the practice of data exploration, instead of enumerating all the topics, a more realistic need is to extract a given number of representative or prominent topics in the data. Based on these representative topics, users can have a quick understanding of the semantic structure of the data collection. Therefore, it is desirable to be able to extract a list of topics ranked according to their importance in the data. Formally, we define the problem as follows:

\begin{definition}
\label{def:diverse_topic_modeling}
\textsl{ Given a data collection $D$, the problem of \textbf{learning top-K topics} aims to infer a list of topics $\{\phi_j\}_{j=1,\cdots K}$ ranked according to their prominence in $D$,  where $K$ is a number specified by the user.}
\end{definition}

Note that different from conventional applications of topic modeling, the number $K$ here is specified completely based on the user's need instead of the prior knowledge of the data, and can be substantially smaller than the true number of topics $\mathcal{K}$ ($K << \mathcal{K}$), which is unknown. To extract top-$K$ representative topics in the data, a straightforward way is to apply a classical topic model (e.g., LDA) directly to fit exactly $K$ topics into the data (i.e., set $\mathcal{K} = K$). In this case, one would expect the inferred topics to be the most representative topics in the data. However, fitting only a few topics to the data usually results in topics that are actually a mixture of multiple topics, the coherence or interpretability of which is compromised. 

As it is difficult for a user to find an appropriate number of topics $\mathcal{K}$, an alternative way is to fit the data with a large number of topics, which usually yields semantically coherent topics. These topics can then be ranked by the proportions of the data they cover (e.g., number of word tokens assigned to each topic). Although these topics are likely to be coherence and meaningful, a large $\mathcal{K}$ may over-fit the data, returning many duplicated topics or smallish topics. As a result, even the top-ranked topics may have a small coverage of the original data, due to the fact that they are either too small or redundant. 
To provide a good summary of the data, it is desirable that the selected topics are both coherent and interpretable, and have a high coverage of the data. Formally, we have

\begin{definition}
\label{def:diverse_topic_modeling}
\textsl{ Given a data collection $D$ and a given integer $K$, the problem of \textbf{topic summarization} aims to find a set of $K$ topics $\{\phi_j\}_{j=1,\cdots K}$ such that every $\phi_j$ is a coherent topic and the set of topics $\{\phi_j\}$ cover as much information in $D$ as possible.}
\end{definition}

Apparently, neither of the simple approaches above is suitable for topic summarization. In next section, we introduce a novel approach, diverse topic modeling, to infer prominent and diverse topics which better summarize the data.


\section{Diverse Topic Modeling}
In most existing topic models or mixture models, the topics or mixture components are generally i.i.d.\ according to a prior distribution. No restriction is placed upon the entire set of topics. This becomes problematic when the data is over-fitted, in which case each underlying theme of original data may be represented by multiple similar topics or be split into topics with a smaller granularity. 

To tackle this problem, the diversity, or redundancy among topics must be considered. Recent work replaces the independent assumption with priors favoring diversified topics (e.g., \cite{DBLP:conf/nips/PetraliaRD12, zou2012priors}), usually through a regularization term over the entire set of topics. A set that contains similar topics will be penalized. Although these treatments can infer more diverse topics than the classical models, they still make the same assumption on the number of topics, and multiple topics are still being used to represent each underlying theme. 

A more reasonable way is to merge similar topics into bigger ones, so that the top-ranked topics become more prominent and also more diverse. In this paper we propose to let the prominent topics merge with or absorb the smaller and similar topics. 
Naturally, we need a process under which topics can communicate with each other and compete for word tokens. If a topic absorbs all the word tokens belonging to another topic, the latter topic is completely absorbed by the former. In the following, we introduce a \emph{reinforced random walk} as such a process. 

\subsection{Reinforced Random Walk}
To model how the topics are competing word tokens from each other, the relationships among the topics must be considered. 
Specifically, we introduce a ``social'' network of topics, or a \emph{topic network}, which is an undirected graph $G=(V, E)$, where $V$ is the set of topics and there is an edge $e(i, j) \in E$ between each pair of topics $(i, j)$. The weight of $e(i, j)$, $w(i, j)$ is defined as the cosine similarity between the word distributions $\phi_i$ and  $\phi_j$, i.e.,
\begin{equation}
w(i,j)=\frac{\phi_i\cdot\phi_j}{||\phi_i||_2\cdot ||\phi_j||_2}.
\end{equation}

We formulate the \emph{absorbing} process among the topics as a random walk process by the word tokens on the topic network. Specifically, if one token belonging to one topic transits to another topic, we can say a token of this topic is \emph{absorbed} by the other topic. If all the tokens belonging to one topic are absorbed by other topics, this topic dies out. 

As we want  the larger topics to gradually absorb the smaller and similar topics, the tokens belonging to the smaller topics should be more likely to transit to the larger ones. This can be achieved through \emph{reinforcing} the transition probability among the topics by the size of the topics, i.e. the number of tokens belonging to the topic. Specifically, the transition probability $p(i, j)$ from topic $i$ to $j$ is defined as 
\begin{equation}
\label{eqn:transition_probability}
	p(i, j)=\frac{p_0(i,j) N_j^{\gamma}}{D_i},
\end{equation}
where $p_0(i,j)$ is the ``organic'' transition probability among the topics, $N_j$ is the size of the topics $j$, $\gamma$ is the parameter used to control the reinforcement intensity by the topic size, and the normalizing factor $D_i$ is calculate as
\begin{equation}
D_i=\sum_jp_0(i,j)N_j^{\gamma}.
\end{equation}

The ``organic'' transition probabilities $p_0(i, j)$ among the topics can be calculated based on the similarity between the topics. Meanwhile, it is also reasonable to assume that each token can also have some probability to stay on the current topic instead of transiting to its neighbors. In this case, tokens belonging to the large topics can still stay on the topic itself. Specifically, we define the $p_0(i, j)$ as follows:
\begin{equation}
p_0(i,j)=\left\{
\begin{array}{c l}      
    \alpha\frac{w(i,j)}{\sum_{j'}w(i,j')}, & \text{if } j \neq i\\
    1-\alpha, & \text{if } j=i
\end{array}\right.,
\end{equation}
where $\alpha$ is the probability of transiting to neighboring topics. 

The above random walk process is related to the stochastic process \emph{vertex-reinforced random walk}~\cite{pemantle1992vertex} and DivRank~\cite{mei2010divrank}, in both of which the transition probabilities among the vertices in the network are reinforced by the number of visits to the vertices. In these processes, the neighboring vertices are actually competing resources from each other.  The reinforcement leads to the \emph{rich-get-richer} phenomena and finally the resources are distributed across the prominent and diverse vertices in the network. 

We embed the above random walk process into the inference processes of  PLSA and LDA, and proposed two diverse topic models DivPLSA and DivLDA. During the inference processes, the topic assignment for each token is first calculated based on the EM algorithm (for PLSA) or Gibbs sampling (for LDA), and then a new topic assignment is obtained based on the reinforced random walk. 

\subsection{Diverse Topic Models}

\subsubsection{DivPLSA}
The PLSA model assumes each document is a mixture of topics with different proportions. Given a collection $D$, the log-likelihood of $D$ under PLSA assumption is calculated as:
\begin{equation}
\label{eqn:plsa}
  L(D)=\sum_d\sum_w n(d,w)\log\sum_{k=1}^\mathcal{K} \theta_{dk}\phi_{kw},
\end{equation}
where $n(d,w)$ is the frequency of word $w$ in document $d$, $\theta_{dk}$ is the probability of topic $k$ in $d$, $\phi_{kw}$ is the probability of word $w$ being generated by topic $k$ and $\mathcal{K}$ is the number of topics specified. The parameters of the model, i.e. $\{\theta_{dk}\}$, and $\{\phi_{kw}\}$,  are estimated by maximizing the log-likelihood. An EM algorithm is generally applied to solve the problem. In the E-step, it calculates the posterior distribution of the topic assignment of each token, i.e. $p(z|d, w)$ based on the current model parameters. In the M-step, it updates the model parameters based on the posterior probability of topic assignments calculated in the E-step. 

We embedded the reinforced random walk process into the E-step. For each token $w$, we first obtain the posterior distribution of topic assignments $p(z|d,w)$ (Eqn.~\eqref{eqn:estep1}). Then one-step reinforced random walk on the topic network is conducted to determine whether this topic is absorbed by other topics or staying on the current topic, which obtains the new topic assignments $p(\hat{z}|d,w)$ (Eqn.~\eqref{eqn:estep2}). The transition probabilities among the topics $P=\{p(i,j)\}$ can be periodically updated in the M-step. We summarize the detailed updating equations as below:

\vspace{1em}
\textbf{E-step:} 
\begin{equation}
\small
\label{eqn:estep1}
  p(z=k|d,w)=\frac{\theta_{dk}\phi_{kw}}{\sum_{k'=1}^K \theta_{dk'}\phi_{k'w}}
\end{equation}
\begin{equation}
\small
\label{eqn:estep2}
 p(\hat{z}=\hat{k}|d,w)=\sum_{k=1}^K p(z=k|d,w)p(k,\hat{k})
 \end{equation}

\textbf{M-step:} 
\begin{equation}
\small
  \theta_{d\hat{k}}\propto \sum_w n(d,w)p(\hat{z}=\hat{k}|d,w)
  \label{eqn:mstep1}
\end{equation}

\begin{equation}
\small
  \phi_{\hat{k}w} \propto \sum_d n(d,w)p(\hat{z}=\hat{k}|d,w)
    \label{eqn:mstep2}
\end{equation}

\begin{equation}
\small
  N_{\hat{k}} = \sum_{d,w} n(d,w)p(\hat{z}=\hat{k}|d,w)
    \label{eqn:mstep3}
\end{equation}

In Alg. 1, we present the detailed learning procedure of the DivPLSA model. Users still need to specify the starting number of topics $\mathcal{K}$, which can be strategically set very large and let the data be over-fitted. During the learning process, the small topics will be eventually absorbed by the larger ones, i.e. the sizes of those topics become 0. We can monitor the number of \emph{active} topics (topics whose sizes are greater than 0) $K^{*}$ during the process. When $K^{*}$ converges, the whole algorithm stops. 

\begin{algorithm}[h]
 \SetAlgoLined
 \SetKwIF{If}{ElseIf}{Else}{if}{then}{else if}{else}{endif} 
 \textbf{Input:} {Training data $D$, the starting number of topics $\mathcal{K}$, the parameter to control the reinforcement intensity by the topic size $\gamma$}\;
 \textbf{Output:} {Number of diverse topics $K^{*}$, the word distributions of topics $\{\phi_k\}_{k=1,\ldots, K^{*}}$ }\;
 \textbf{initialization:} randomly initialize the document-topic and topic-word distributions\;
 calculate the sizes of the topics and the transition probabilities among the topics according to \eqref{eqn:transition_probability}\;
 \While{$K^{*}$ no convergence}{
    \textbf{E-step:} for each word $w$ of each document $d$ in $\{1,\ldots,|D|\}$
    \begin{itemize}
    	\item calculate the posterior probability of topic assignments $p(z|d,w)$  according to \eqref{eqn:estep1}\;
	\item conduct one-step reinforced random walk on \\the topic network according to \eqref{eqn:estep2}\;
    \end{itemize}
    \textbf{M-step:} 
    \begin{itemize}
    	\item update document-topic, topic-word distributions  \\and topic sizes according to \eqref{eqn:mstep1}, \eqref{eqn:mstep2} and \eqref{eqn:mstep3}\;
   	\item update the transition probability \\among the topics according to \eqref{eqn:transition_probability}\;
	\item calculate the number of active topics $K^{*}$\;
   \end{itemize}
 }
 \caption{The DivPLSA model}
\end{algorithm}

\subsubsection{DivLDA}
The LDA model is a Bayesian treatment of the PLSA by placing the Dirichlet priors upon both the document-topic and topic-word distributions, which results the model is computationally intractable.  Collapsed Gibbs sampling~\cite{griffiths2004finding} algorithm is widely used for the inference due to its simplicity and effectiveness. In Gibbs sampling, each dimension of the joint distribution is sampled alternatively based on the distribution conditioned on all the other variables. Specifically, in LDA the conditional distribution of the topic assignment $z_{i}$ associated with word $w_i$ is calculated via:
\begin{equation}
\label{eqn:lda}
  p(z_i=k|w_i=w,\mathbf{w}_{-i},\mathbf{z}_{-i}) \propto (n_{d,k}+\alpha_k)\frac{n_{k,w}+\beta}{n_k+V\beta}.
\end{equation}
$\mathbf{z}_{-i}$ represents the topic assignments of all the words excluding the current word. $n_{d,k}$ is the number of words assigned to the $k^{th}$ topic in document $d$; $n_{k,w}$ is the number of times that the word $w$ is assigned to the $k^{th}$ topic; and $n_k$ is the number of tokens assigned to the $k^{th}$ topic, i.e., the size of topic $k$. For all these counts, the current token is excluded.

In DivLDA, instead of calculating the expectation of the topic assignments as done in the DivPLSA, sampling is used to obtain the topic assignments. For each token, we first sample the topic assignment $z_i$ based on equation~\eqref{eqn:lda}, and then conduct one-step reinforced random walk starting from $z_i$ to obtain a new topic assignment $\hat{z}_i$ according to 
\begin{equation}
\label{eqn:lda2}
  p(\hat{z}_i=\hat{k}|z_i=k)=p(k,\hat{k})
\end{equation}

The final document-topic and topic-word distributions can be estimated via:

\begin{equation}
\theta_{dk}=\frac{n_{dk}+\alpha_k}{\sum_{k'}(n_{dk'}+\alpha_{k'})}
\end{equation}

\begin{equation}
\phi_{kw}=\frac{n_{kw}+\beta}{n_k+V\beta}
\end{equation}

The detailed inference process is summarized in Alg. 2. 

\begin{algorithm}[h]
 \SetAlgoLined
 \SetKwIF{If}{ElseIf}{Else}{if}{then}{else if}{else}{endif} 
 \textbf{Input:} {Training data $D$, the starting number of topics $\mathcal{K}$,  the parameter to control the reinforcement intensity by the topic size $\gamma$}, the total number of Gibbs sampling iterations $M$\; 
 \textbf{Output:} {Number of diverse topics $K^{*}$, the word distributions of topics $\{\phi_k\}_{k=1,\ldots,K^{*}}$ }\;
 \textbf{initialization:} randomly sample the topic assignments for all the word tokens.\;
 calculate the sizes of the topics and the transition probabilities among the topics according to \eqref{eqn:transition_probability}\;
 \While{ iter $\leq$ M}{
    for each token $w$ of each document $d$ in $\{1,\ldots,|D|\}$
     \begin{itemize}
     	\item for the current assignment $\tilde{k}$ of token $w$, decrement counts and sums: $--n_{d\tilde{k}}$, $--n_{\tilde{k}w}$, $--n_{\tilde{k}}$\;
	\item sample the topic assignments $z_i=k$ for the token  according to~\eqref{eqn:lda}\;
	\item conduct one-step reinforced random walk on the topic network according to~\eqref{eqn:lda2}\;
	\item for the latest assignment $\hat{k}$, increment counts and sums: $++n_{d\hat{k}}$, $++n_{\hat{k}w}$,$++n_{\hat{k}}$\;
     \end{itemize}
     update the transition probability among the topics according to \eqref{eqn:transition_probability}\;
     optimize the parameter $\vec{\alpha}$ according to~\cite{blei2003latent}\;
 }
 \caption{The DivLDA model}
\end{algorithm}

\subsection{Discussion}
We discuss some practical issues of the two models. 

\noindent\textbf{Convergence.}
In the above two models, after plugging in the reinforced random walk process within the EM or Gibbs sampling process, there are no explicit objective functions any more. We empirically prove that both the number of active topics and the data likelihood will converge (see Figure~\ref{fig:divplsa-topic-num} and~\ref{fig:divplsa-likelihood}). We leave the theoretical justification of the convergence as the future work.

\noindent\textbf{How to set the parameter $\alpha$?} 
The parameter $\alpha$ controls the transition probability to the neighbor topics or staying on itself. It takes similar effect as the step size in the gradient descent method, and hence in practice a small $\alpha$ (e.g., 0.1) can be used. 

\noindent\textbf{How to set the parameters $\gamma$ and $\mathcal{K}$?}
We empirically show that the performance of the two models are not sensitive to the parameters $\gamma$ and $\mathcal{K}$ (See Figure~\ref{fig:ps-gamma-20ng} and~\ref{fig:ps-k-20ng}). In practice, $\mathcal{K}$ can be set to be very large to let the data over-fitted. $\gamma$ can be usually set within $[1, 2]$ for DivPLSA and $[0.6,1.5]$ for DivLDA. 

\noindent\textbf{Scalability.}
Both DivPLSA and DivLDA can be easily scaled by making use of existing large-scale topic modeling techniques. The E-step of DivPLSA can be parallelized by assigning the documents to different processors or nodes. A scaled version of DivLDA can be built on top of the existing large scale LDA model, e.g., the yahoo-LDA model in~\cite{ahmed2012scalable}.

\section{Experiment}
In this section, we move forward to evaluate the effectiveness of our proposed diverse topic models. We evaluate the performances on four real-world data sets. 
\subsection{Datasets}
\noindent \textbf{\textsc{4CONF.}}  We start with a small data set, the 4CONF data set as in~\cite{mei2008topic}. The data set is constructed from papers published in four conferences including KDD, SIGIR, NIPS, and WWW. Every document corresponds to an author by aggregating the titles of the author's papers. Stop words and words appearing in less than 10 documents are removed. This small data set allows us to interpret the topics intuitively and visually. 
 
\noindent \textbf{\textsc{20NG.}} This is the widely used 20 newsgroup data set in text mining. Stop words and words appearing in less than 20 documents are removed. We sample 1,000 documents from the set as a holdout data set. 

\noindent \textbf{\textsc{WIKIPEDIA.}} This includes 10,000 articles randomly sampled from 4,636,797 Wikipedia articles in English. Stop words and words appearing in less than 100 documents are removed. We also hold out a sample of 1,000 articles.  
 
\noindent \textbf{\textsc{DBLP.}} The larger data set consists of all papers with abstracts in the computer science bibliography as in~\cite{Tang:08KDD}. Stop words and words appearing in less than 50 documents are removed.
    
Table 1 summarizes the statistics of all the data sets. 
  
\begin{table}[htbp]
\caption{Statistics of the data sets}
\label{tab:dataset-statistics}
\begin{center}
\scalebox{0.9}{
\begin{tabular}{|c|c|c|c|c|c|} \hline
Dataset & \# train & \# holdout& vocabulary & \# tokens \\ \hline\hline
4CONF &8,486&--&1,672&80,642 \\ \hline
20NG &11,267&1,000&7,642&1,056,012 \\ \hline
WIKIPEDIA &10,000&1,000&196,665&3,006,817 \\ \hline
DBLP &529,434&--&25,404& 36,899,908\\ \hline
\end{tabular}
}
\end{center}
\end{table}

\subsection{Evaluation Metrics}
To provide a good summary of the data, the topics need to be highly coherent and also provide a high coverage of the information in the original data. We introduce metrics to evaluate the semantic coherence and the coverage of the topics respectively. The quality of a topic is measured through the \emph{semantic coherence} of the word distribution while the information coverage of a set of topics is measured through their predictive performance on the holdout data set, using the well-adopted \emph{perplexity} metric.

\vspace{1em}\noindent \textbf{Topic Semantic Coherence.} We measure the semantic quality of the topics through the semantic coherence of the topics. In~\cite{newman2010automatic}, Newman et al.\  measures the semantic coherence of each topic as the average point-wise mutual information (PMI) of every word pair among the top-ranked words in the topic. Specifically, the overall semantic topic coherence of a set of topics $\Phi=\{\phi_k\}_{k=1}^K$ is calculated as:
\begin{equation}
\small
  \text{PMI}(\Phi)=\frac{1}{K}\sum_{k=1}^K\frac{2}{N(N-1)}\sum_{1 \leq i<j \leq N}\log \frac{p(w_{ki},w_{kj})}{p(w_{ki})p(w_{kj})},
\end{equation}
where $w_k=(w_{k1},\ldots, w_{kN})$ are the words ranked at the top $N$ positions in topic $\phi_k$. $p(w_{ki}, w_{ki})$ is the probability that the pair of words co-occur in the same document while $p(w_{ki})$ is probability of a word $w_{ki}$ appearing in a single document. The top ranked 20 ($N=20$) words are used in our experiments. 

To calculate the PMI, generally a large dataset has to be used. The entire 20NG data set (12,267 documents) and the entire English Wikipedia (4.6 million documents) are used in calculating the PMI for the experiments on the 20NG data set and on the WIKIPEDIA data set, respectively. 

\vspace{1em}\noindent \textbf{Perplexity.} We measure the information coverage of a set of topics by \emph{perplexity}, which measures the predictive performance of these topics on the holdout data set. Specifically, each document $w_i$ in the held-out data set is split into two parts $w_i=(w_{i1},w_{i2})$. The likelihood of the second part of the documents $w_{i2}$ ($20\%$ of the document) is calculated based on the training data $D$ and the first $80\%$ words of the document $w_{i1}$. Specifically, we have 

\begin{equation}
\small
  Perplexity=\exp\big\{-\frac{\sum_i\log p(\textbf{w}_{i2}|\textbf{w}_{i1}, D_{\text{train}})}{\sum_i{|\textbf{w}_{i2}|}}\big\}
\end{equation}

\subsection{Algorithms for Comparison}
We compare the following algorithms for selecting \emph{top-K} topics for data summarization.
\begin{itemize}
	\item \textbf{PLSA/LDA.} The classical PLSA or LDA model is directly utilized to learn exactly \emph{K} topics.
	\item \textbf{PLSA/LDA-TopK.} We first train PLSA/LDA with a large number of $\mathcal{K}$ topics, and then pick \emph{top-K} topics with the largest sizes. 
	\item \textbf{PLSA/LDA-MMR-TopK.} PLSA/LDA is used to train a large number of $\mathcal{K}$ topics. Then the MMR algorithm~\cite{carbonell1998use} is used to select \emph{top-K} topics from the $\mathcal{K}$ topics. Although the MMR algorithm is proposed in a query-dependent setting, we adapt it to our scenario. The relevance between the query and each topic is measured as the coverage of this topic in the whole data set, and the similarity among each pair of topics is calculated as the cosine similarity of the word distributions. The best results are reported by empirically tuning the parameter $\lambda$ in the MMR algorithm.
	\item \textbf{PLSA/LDA-DivRank-TopK.} PLSA/LDA is used to train a large number of $\mathcal{K}$ topics, and then the DivRank algorithm~\cite{mei2010divrank} is used to select \emph{top-K} topics. In the DivRank algorithm, we treat the proportions of the topics as the preference vector, and the weight between each pair of topics is calculated as the cosine similarity of the corresponding word distributions. The best results are reported by empirically tuning the parameter $\alpha$ and $\lambda$ in the DivRank algorithm.
	\item \textbf{DivPLSA/DivLDA-TopK.} DivPLSA/DivLDA is applied on the data set and then the \emph{top-K} topics with the largest sizes are selected.
\end{itemize}

\begin{figure*}[htdb!]
\centering
\subfigure[Iteration 1]{
\includegraphics[width=1.5in]{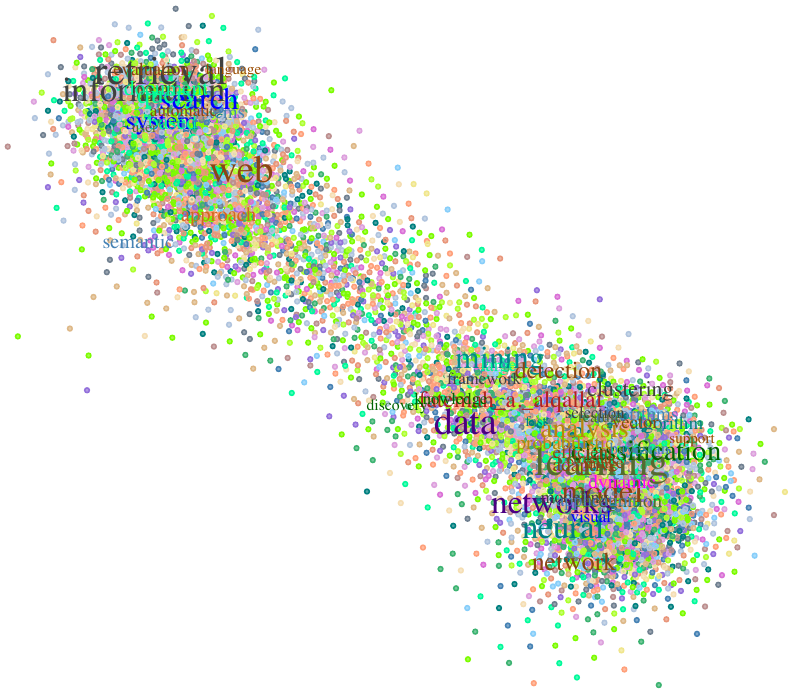}
\label{fig:divplsa-1}
}
\subfigure[Iteration 20]{
\includegraphics[width=1.5in]{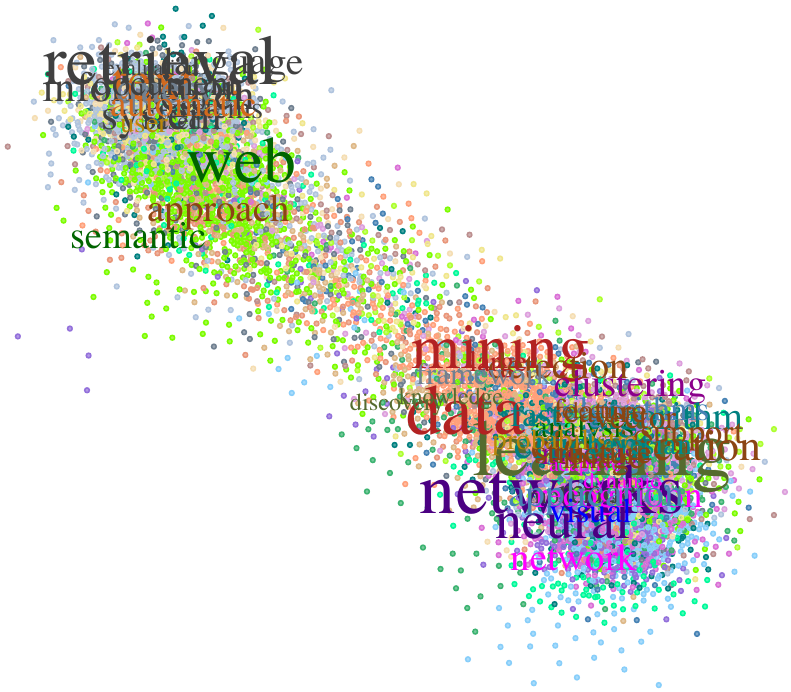}
\label{fig:divplsa-20}
}
\subfigure[Iteration 50]{
\includegraphics[width=1.5in]{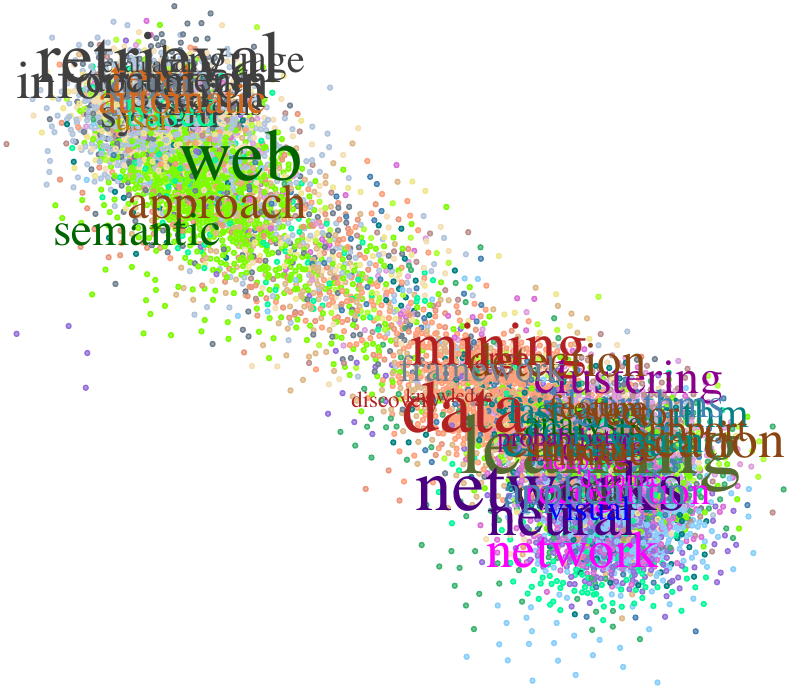}
\label{fig:divplsa-50}
}
\subfigure[\#topics v.s. \#iterations]{
\includegraphics[width=1.5in]{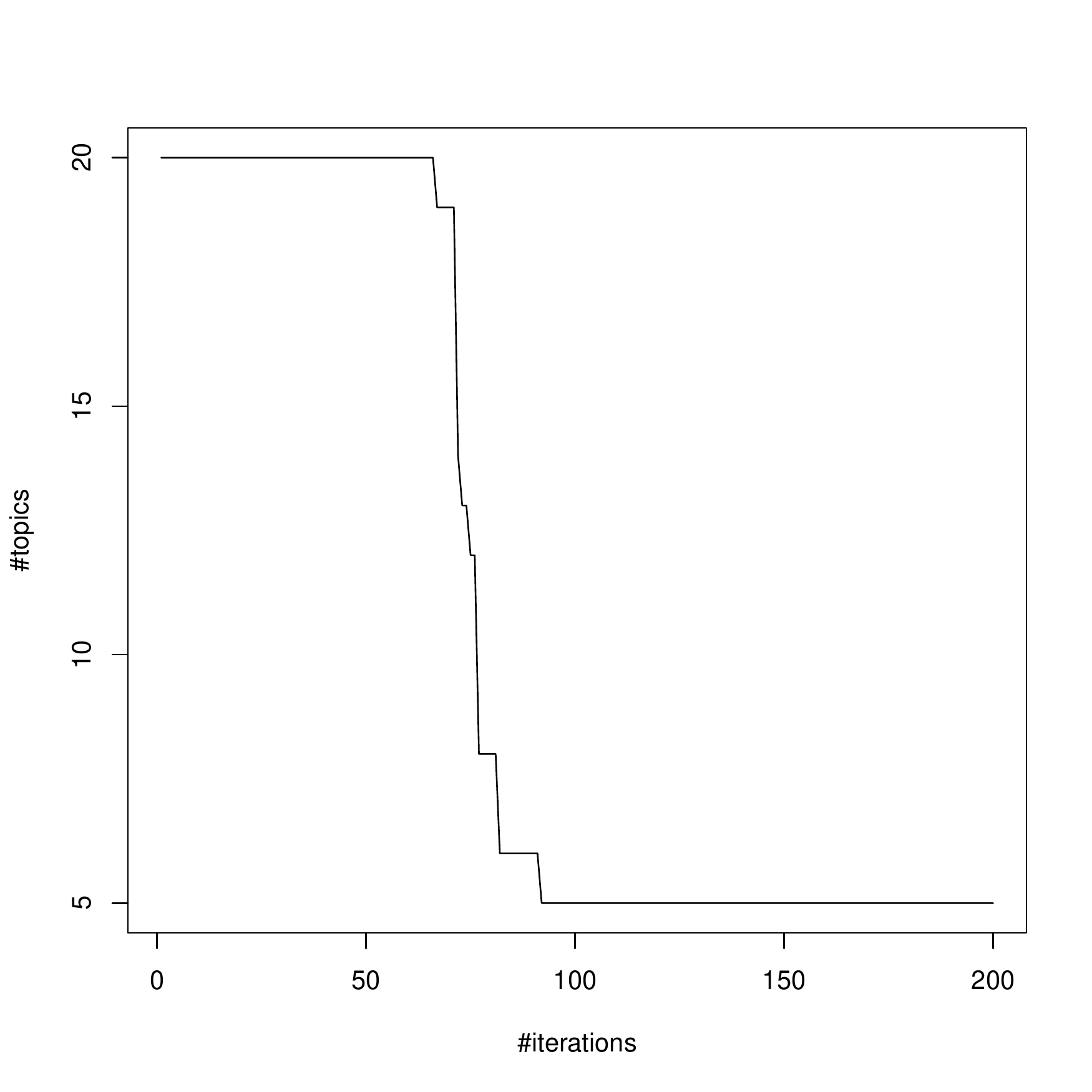}
\label{fig:divplsa-topic-num}
}
\subfigure[Iteration 70]{
\includegraphics[width=1.5in]{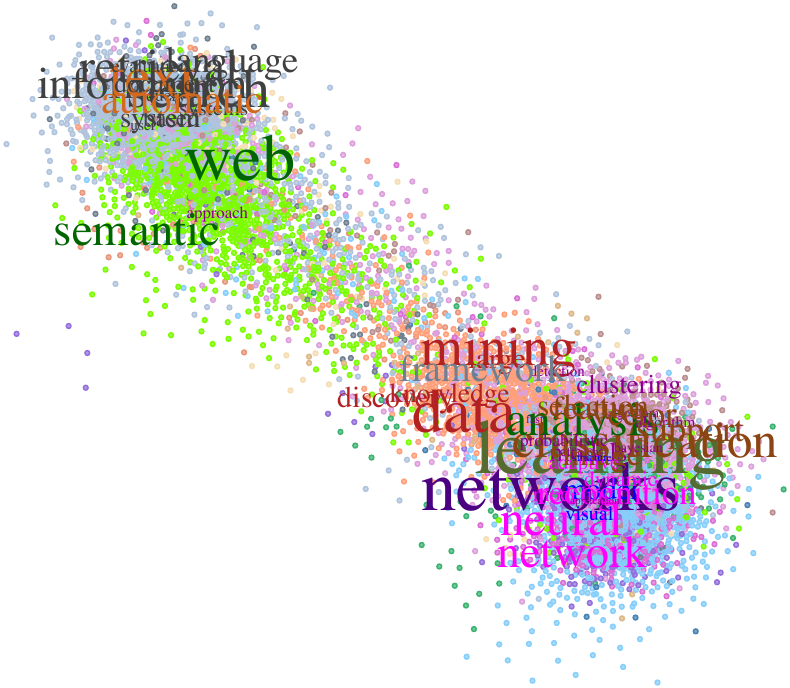}
\label{fig:divplsa-70}
}
\subfigure[Iteration 90]{
\includegraphics[width=1.5in]{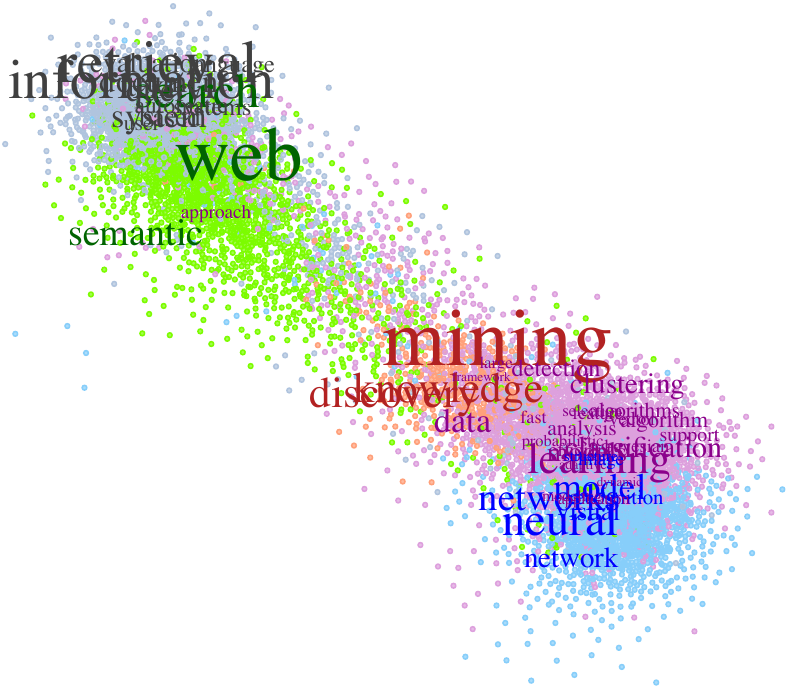}
\label{fig:divplsa-90}
}
\subfigure[Iteration 200]{
\includegraphics[width=1.5in]{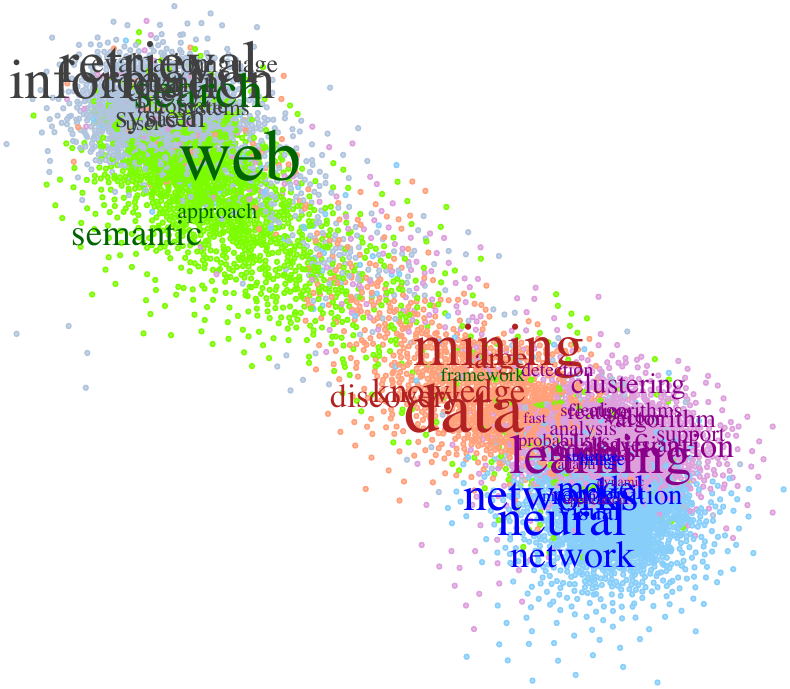}
\label{fig:divplsa-200}
}
\subfigure[likelihood v.s.\ \#iterations]{
\includegraphics[width=1.5in]{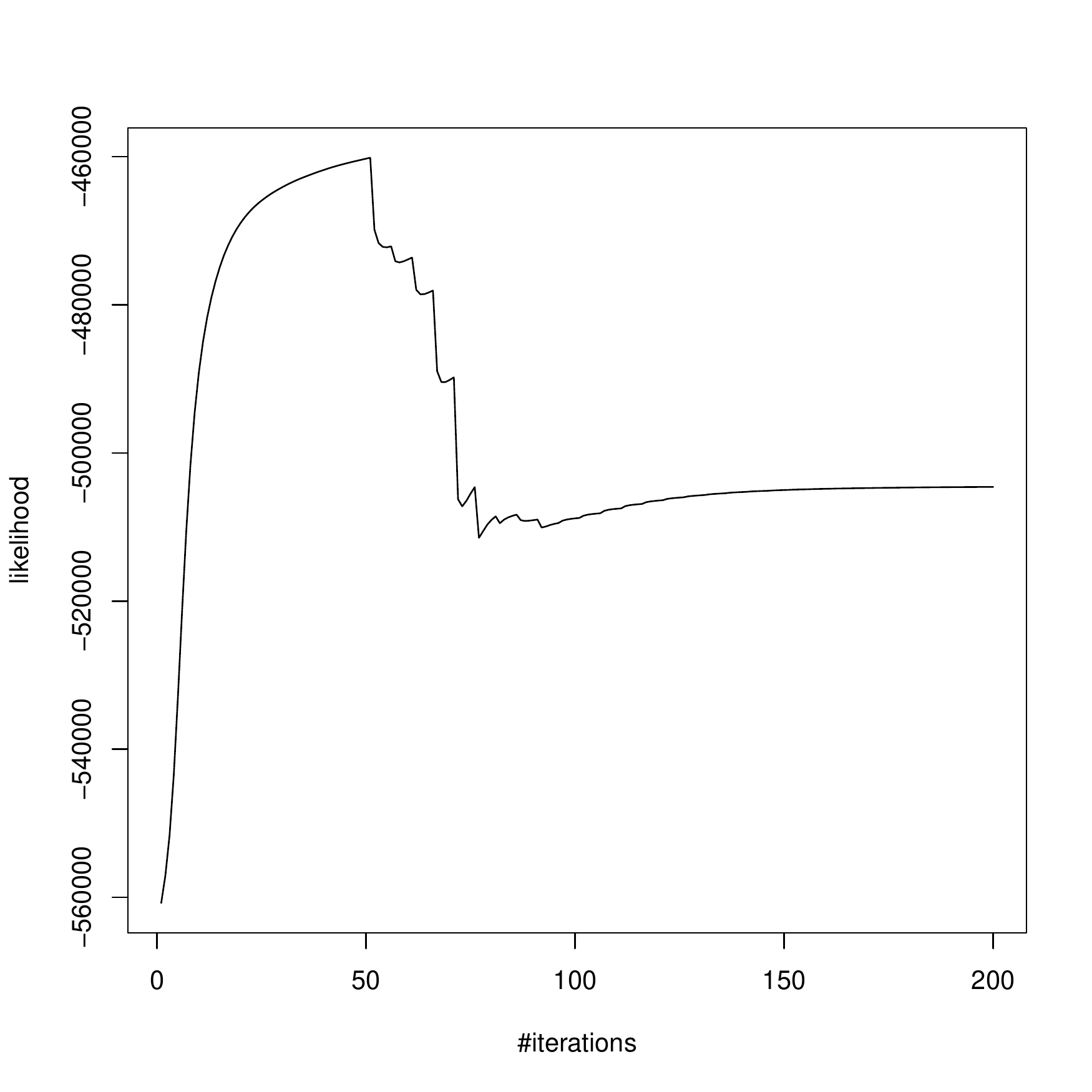}
\label{fig:divplsa-likelihood}
}
\caption{The learning behavior of DivPLSA in 4CONF dataset.}
\label{fig:divplsa-animation}
\end{figure*}

\subsection{Learning Behavior of Diverse Topic Models}
We start the experimental results by investigating the learning behavior of our proposed diverse topic models. We take the DivPLSA model as an example. Similar behavior of the DivLDA is observed and hence we do not include here. In Figure~\ref{fig:divplsa-animation}, we show how the topic assignments of the documents and words change along the iterations in the 4CONF dataset. We first build a document-word bipartite network with the edge weight as the word frequency in the document. A DrL~\cite{martin2008drl} layout algorithm is applied on this bipartite network to calculate the two-dimensional coordinates for both the documents and words. In Figure~\ref{fig:divplsa-1}, each data point represents a document, and the most popular 100 words in the dataset are shown. Different colors represent different topics, and each document or word is assigned to its most probable topic. The size of each word is determined based on the probability of the word in the assigned topic. We expect that documents and words belonging to the same topics (colors) are likely to lie within a dense area.

We start with 20 topics in the DivPLSA model with $\alpha=0.1, \gamma=1.9$. In the first 50 EM iterations, the same EM iteration for PLSA is conducted and the reinforced random walk process  kicks in after the 50th iteration. In the first EM iteration (Figure~\ref{fig:divplsa-1}), the topic assignments of all the tokens in the data set are randomly initialized and we can see that the colors in the entire plot are totally mixed. There is no dense area taking the same color. As time goes on, dense area with the same color gradually emerges, e.g., the 20th iteration (Figure~\ref{fig:divplsa-20}). The results become better when reaching the 50th iteration. We can see the emergence of topics such as ``information retrieval'', ``Web semantic''. However, we can see that too many topics are fitted into the data, and many of them are small and similar to each other. No clear topic semantic structure is discovered. After the 50th iteration, the reinforced random walk process kicks in and the smaller topics are gradually absorbed by the larger and similar ones. We can see better topical structure in 70th iterations, in which fewer than 20 topics are active. In the 200th iteration, the whole procedure stops and five active topics remain (the other 15 topics are fully absorbed) including ``information retrieval", ``web'', ``data mining'', ``neural network'' and ``learning, Bayesian''. Recall that the dataset is collected from the four conferences ``SIGIR'', ``WWW'', ``KDD'' and ``NIPS'', the five topics are a good summary of the original data. It is also interesting to notice that two diverse topics ``neural network" and ``learning, Bayesian'' are discovered in the ``NIPS'' conference, which indicates there are two distinctive research communities in the machine learning area. The final list of topics is represented in Table~\ref{tab::DivPLSA}.

\begin{table}[htbp]
\caption{Topics extracted by DivPLSA \label{tab::DivPLSA}} 
\scriptsize
\begin{tabular}{c} \hline
\textsc{Top-ranked Words of Each Topic} \\ \hline \hline
learning classification models clustering algorithm support bayesian  \\\hline
retrieval information text query system document evaluation \\\hline
neural networks network model recognition visual learning \\ \hline
web search semantic xml extraction analysis content \\ \hline
data mining knowledge discovery large databases rules \\ \hline
\end{tabular}
\end{table}

In Figure~\ref{fig:divplsa-topic-num}, we show how the number of active topics changes along the iterations. In the first 50 iterations, the number of active topics remains 20 as the random walk process has not kicked in yet.  The number of active topics begins  to decrease in around the 70th iterations as the smaller topics are absorbed by the larger and similar ones, and dramatically decreases to 5 in around the 90th iterations. The number of active topics converges after 90 iterations. During the process, we can see that the absorbing process converges quite quickly, taking around 20 EM iterations.

Figure~\ref{fig:divplsa-likelihood} shows how the likelihood of the training data changes over iterations. The likelihood keeps increasing at first and starts to decrease in around the 70th iteration, as the number of active topics decreases. In around the 90 iterations, the likelihood stops decreasing as the number of active topics converges. After this, the likelihood keeps increasing and converges in around the 150th iterations. 


Overall, we can see that when the data is over-fitted, the inferred topics by the classical topic models tend to be small and duplicated, which is not good for data summarization purpose. The reinforced random walk is able to merge the similar topics and end up with some large diverse topics. Both the number of topics and the likelihood of the data will finally converge. 
\begin{figure*}[!htdb]
    \begin{minipage}{0.49\linewidth}
\centering
\subfigure[Perplexity]{
\includegraphics[width=1.5in]{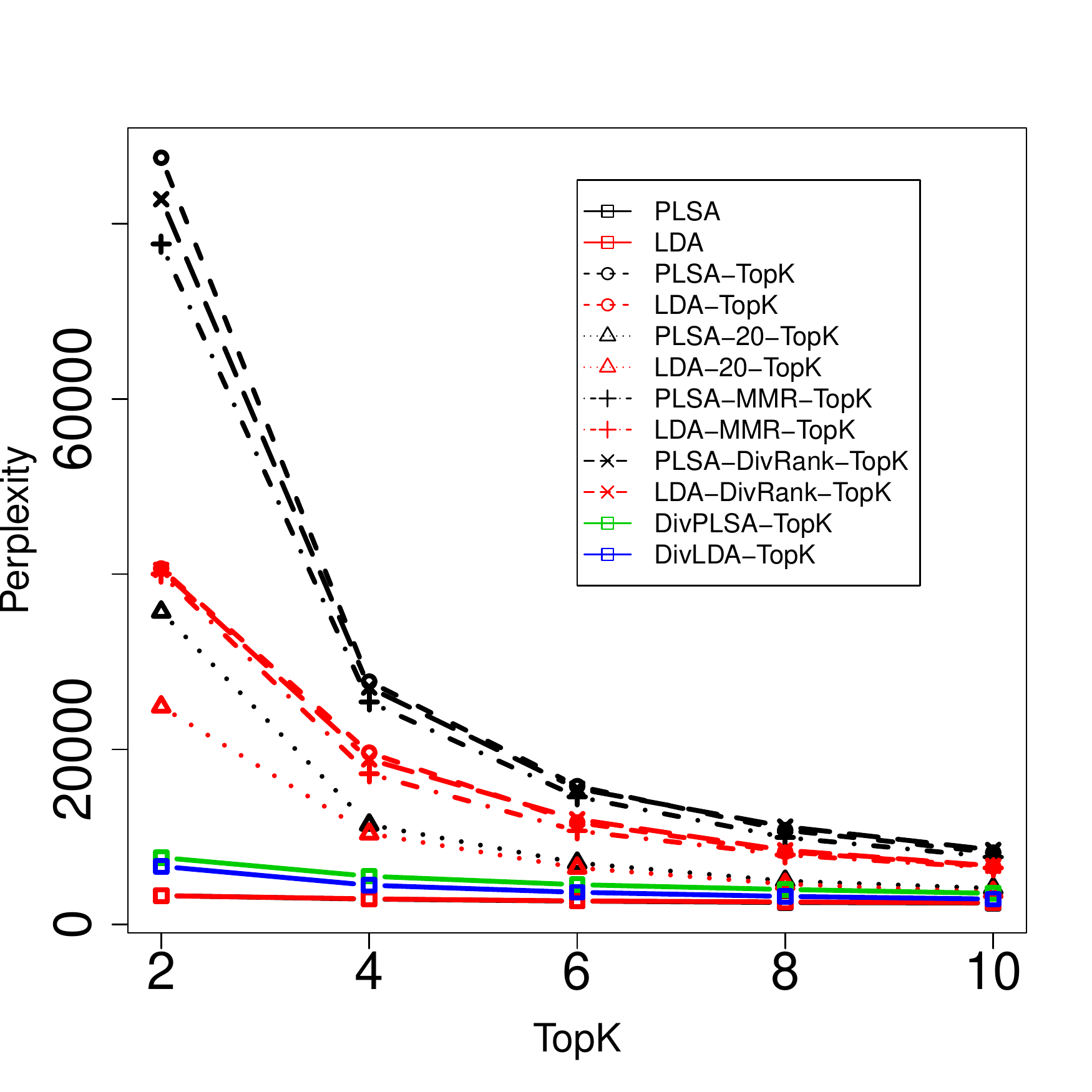}
\label{fig:summarization-20ng-perplexity}
}
\subfigure[Semantic Coherence]{
\includegraphics[width=1.5in]{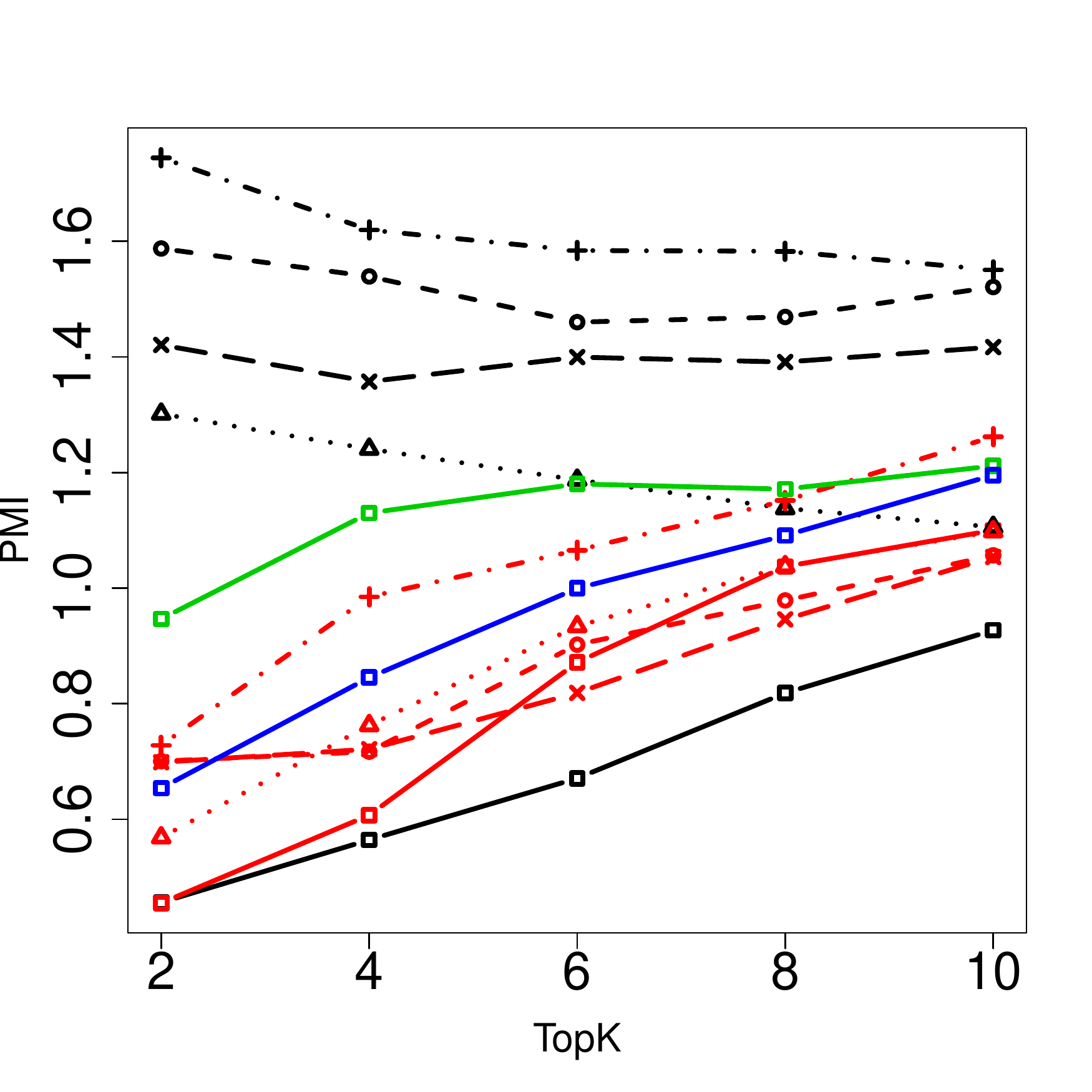}
\label{fig:summarization-20ng-pmi}
}
    \end{minipage}
    \begin{minipage}{0.49\linewidth}
\centering
\subfigure[Comparison of the topics learned by different methods. ]{
\scalebox{0.7}{
\label{fig:summarization-20ng-topics}
\begin{tabular}{|c|c|c|c|} \hline
Algorithm&Top-2 Topics& PMI&Proportion \\ \hline\hline
\multirow{2}{0.5in}{PLSA}&people writes article don god time good apr&0.499&58.63\%\\
&writes system article file mail don windows key&0.449&41.37\%\\ \hline	
\multirow{2}{0.5in}{PLSA-TopK}&god jesus church christ bible lord man faith&1.845	&2.98\%\\	
& server mit sun motif source version library tar&1.905	&2.95\%\\	\hline
\multirow{2}{0.5in}{DivPLSA-TopK} & god people jesus christian bible don life time& 1.177&12.48\%\\ 
 &file program window files ftp image version server&1.325 &12.19\%\\ \hline
\end{tabular}}
}
    \end{minipage}
\caption{Performances on the 20NG dataset. }
    \label{summarization-20NG}
\end{figure*}

\begin{figure*}[!htdb]
    \begin{minipage}{0.49\linewidth}
\centering
\subfigure[Perplexity]{
\includegraphics[width=1.5in]{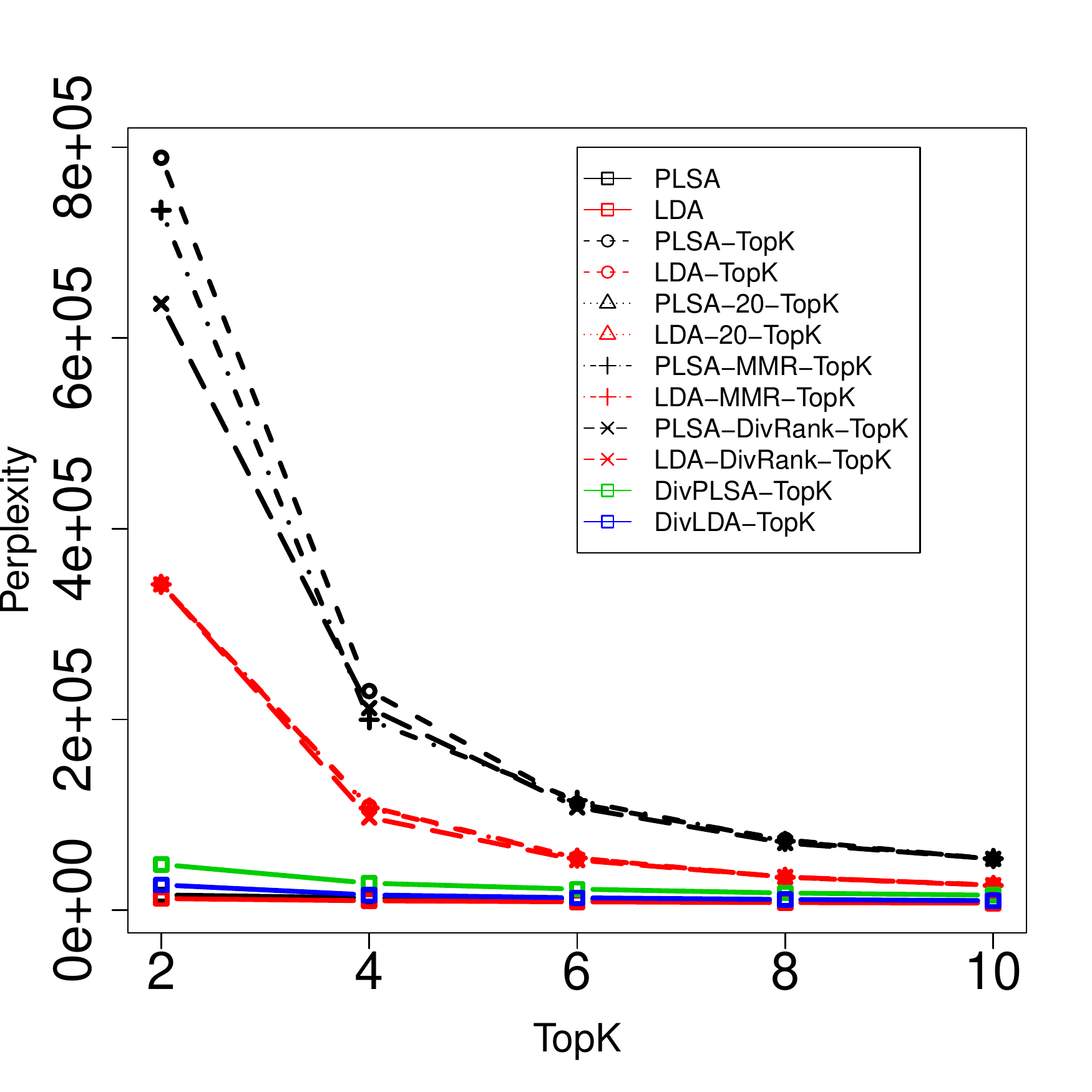}
\label{fig:summarization-wikipedia-perplexity}
}
\subfigure[Semantic Coherence]{
\includegraphics[width=1.5in]{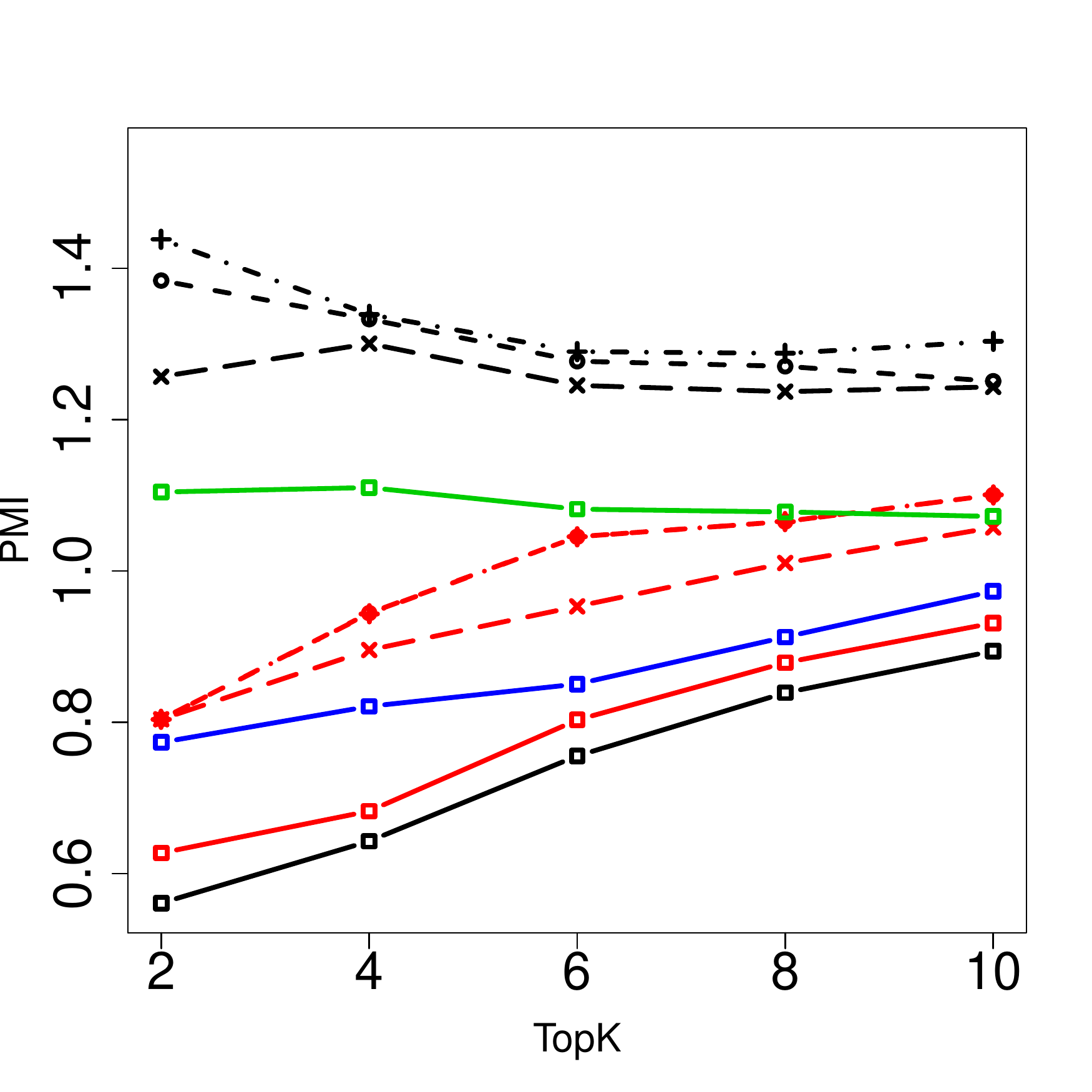}
\label{fig:summarization-wikipedia-pmi}
}
    \end{minipage}
    \begin{minipage}{0.49\linewidth}
\centering
\subfigure[Comparison of the topics learned by different methods. ]{
\scalebox{0.7}{
\label{fig:summarization-wikipedia-topics}
\begin{tabular}{|c|c|c|c|} \hline
Algorithm&Top-2 Topics& PMI&Proportion \\ \hline\hline
\multirow{2}{0.5in}{PLSA}& user overlap university world school calculated time war &0.295&52.70\%\\
&template article page wikipedia made delete image user&0.862&47.30\%\\ \hline	
\multirow{2}{0.5in}{PLSA-TopK}&album music song band released rock records single &1.590	&2.57\%\\	
& wikipedia http php user page article talk edit&	1.280&2.40\%\\	\hline
\multirow{2}{0.5in}{DivPLSA-TopK} & film series show episode time man television movie&1.051 &8.51\%\\ 
 &music album song band released single rock songs & 1.563&5.23\%\\ \hline
\end{tabular}}
}
    \end{minipage}
\caption{Performances on the WIKIPEDIA dataset. }
    \label{fig:summarization-wikipedia}
\end{figure*}

\subsection{Evaluation for Data Summarization}
Next, we move forward to evaluate the performances of different algorithms for the task of topic summarization, which aims to learn and select \emph{top-K} topics for summarizing the data. The performances are evaluated in terms of the semantic coherence and information coverage of the topics, which are measured by PMI and perplexity respectively. 

The results on the 20NG dataset are presented in Figure \ref{summarization-20NG}. Figure \ref{fig:summarization-20ng-perplexity} presents the information coverage of the topics learned by various algorithms in terms of perplexity. The lower the perplexity, the larger the information coverage. Overall, the more topics are selected, the higher the information coverage. All the methods except PLSA/LDA are trained with 50 topics (the starting number of topics is also 50 for DivPLSA/DivLDA).
The PLSA-TopK algorithm, which selects the largest \emph{top-K} topics in the 50 topics trained by the PLSA model, achieves the worst performance. In this case, the data is over-fitted and the inferred topics are small and similar to each other, which would have a low information coverage of the data. Both PLSA-MMR-TopK and PLSA-DivRank-TopK outperform PLSA-TopK by selecting the non-redundant largest \emph{top-K} topics. 
 LDA-TopK achieves better performance than PLSA-TopK, which may due to the fact that the largest topics learned by LDA is larger than the ones learned by PLSA through an examination of the sizes of the largest topics learned by the two models. Similarly, LDA-MMR-TopK and LDA-DivRank-TopK further improve the information coverage by taking into the diversity among the selected topics into consideration. 
 
 As there are 20 categories in the 20NG dataset, it is reasonable to think there are 20 topics in the dataset. Therefore, we trained PLSA/LDA with 20 topics and the topics with the largest size are selected (marked as PLSA/LDA-20-TopK). We can see that the topics selected from 20 topics (PLSA/LDA-20-TopK) has a high information coverage than from 50 topics (PLSA/LDA-TopK). This indicates that if we can have a good estimation of the appropriate number of topics in the data, it is likely to infer better topics with a high information coverage. Our DivPLSA and DivLDA models (also starting from 50 topics) outperform all these models by learning the most prominent and diverse topics without worrying about estimating the appropriate number of topics. Though starting from a large number of topics, the DivPLSA/DivLDA model is able to merge those similar topics and end up with the diverse ones. We are also surprised to see that deploying PLSA/LDA to train just $K$ topics (denoted as PLSA/LDA) obtains the best performance in terms of information coverage. However, though the discovered topics have a high information coverage in terms of perplexity, the semantic coherence of the topics is pretty low and the topics are not interpretable to users.

In Figure \ref{fig:summarization-20ng-pmi}, we compare the semantic coherence of the topics inferred by different algorithms. Though the $K$ topics directly learned by PLSA/LDA  have a high information coverage, the semantic coherence of the topics is the worst. This is because the data is under-fitted and the inferred topics are a mixture of multiple topics. Therefore, the topics  are more likely to be the background topic of the whole data collection (See the first row in Figure \ref{fig:summarization-20ng-topics}) and do not have any semantic information. The semantic quality of the \emph{top-K} topics selected from 50 topics learned by PLSA using various ways (PLSA-TopK, PLSA-MMR-TopK, PLSA-DivRank-TopK) are the best among all the models. This may due to the granularity of these topics is quite small\footnote{Topics with a small granularity tend to yield a larger PMI}, which can be examined from the second row in Figure \ref{fig:summarization-20ng-topics}. The topics learned by the DivPLSA and DivLDA have a reasonable good semantic quality meanwhile have a high information coverage, both of which are desirable for a good summarization. 

Figure \ref{fig:summarization-wikipedia} compares the performance on the WIKIPEDIA data set. To select an appropriate number of topics within the data, we vary different numbers of topics and choose the best one 100 based on the predictive performance on the holdout data set. All the models (except PLSA and LDA directly trained with a few topics) are trained with 100 topics. Similar behavior of different models can be observed in the data set as in the 20NG data set. 

To summarize, our DivPLSA and DivLDA models give a good summarization of the data by presenting \emph{top-K} topics with a high information coverage and a reasonable good semantic quality.  The \emph{K} topics directly trained with PLSA/LDA are not interpretable and cannot convey useful semantic information to users at all.  The \emph{top-K} topics selected from a large number of topics trained by PLSA/LDA have good semantic quality but low information coverage. 

\subsection{Parameter Sensitivity}
\begin{figure*}[htdb!]
\centering
\subfigure[Perplexity (DivPLSA)]{
\includegraphics[width=1.5in]{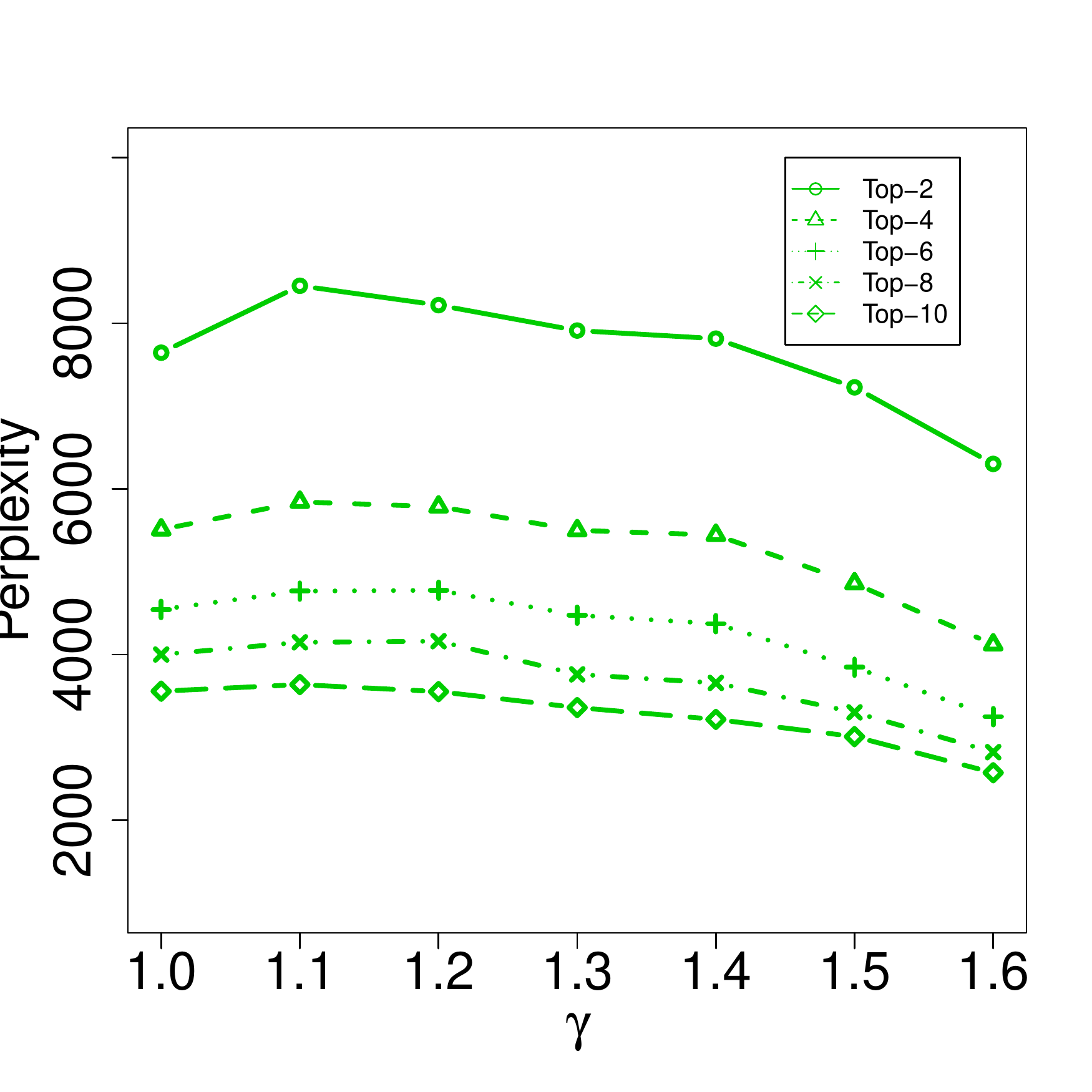}
\label{fig:subfig1}
}
\subfigure[Semantic Coherence (DivPLSA)]{
\includegraphics[width=1.5in]{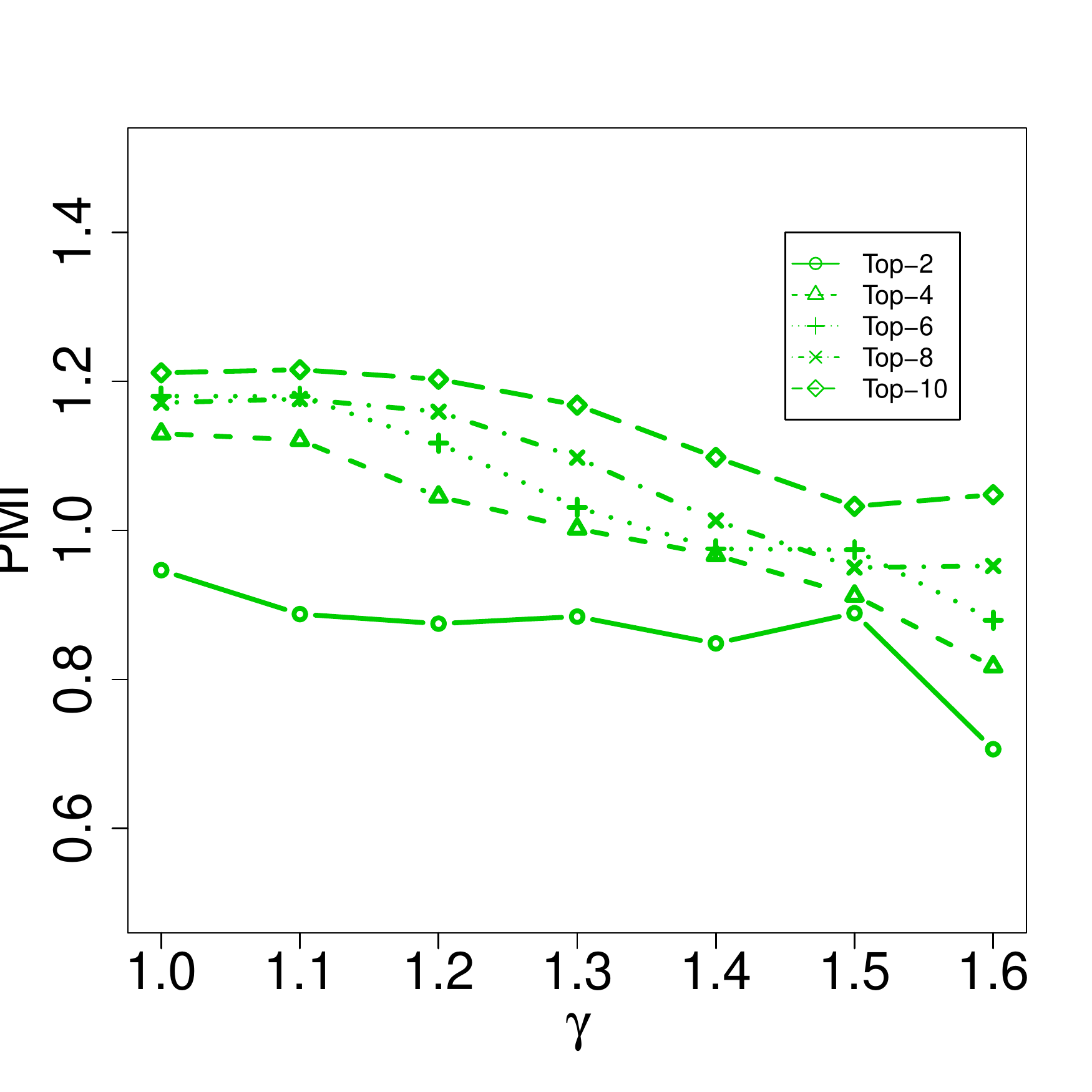}
\label{fig:subfig1}
}
\subfigure[Perplexity (DivLDA)]{
\includegraphics[width=1.5in]{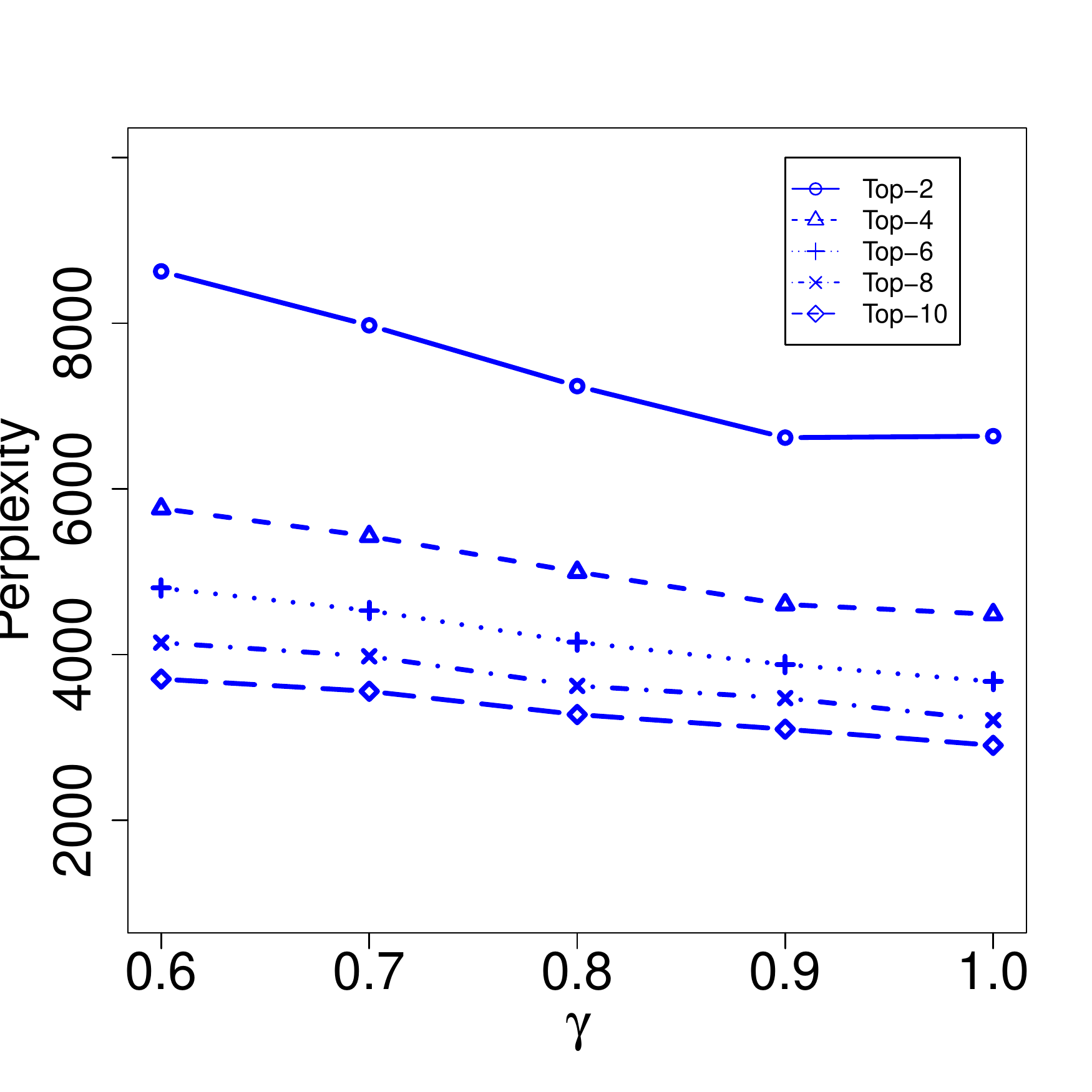}
\label{fig:subfig2}
}
\subfigure[Semantic Coherence (DivLDA)]{
\includegraphics[width=1.5in]{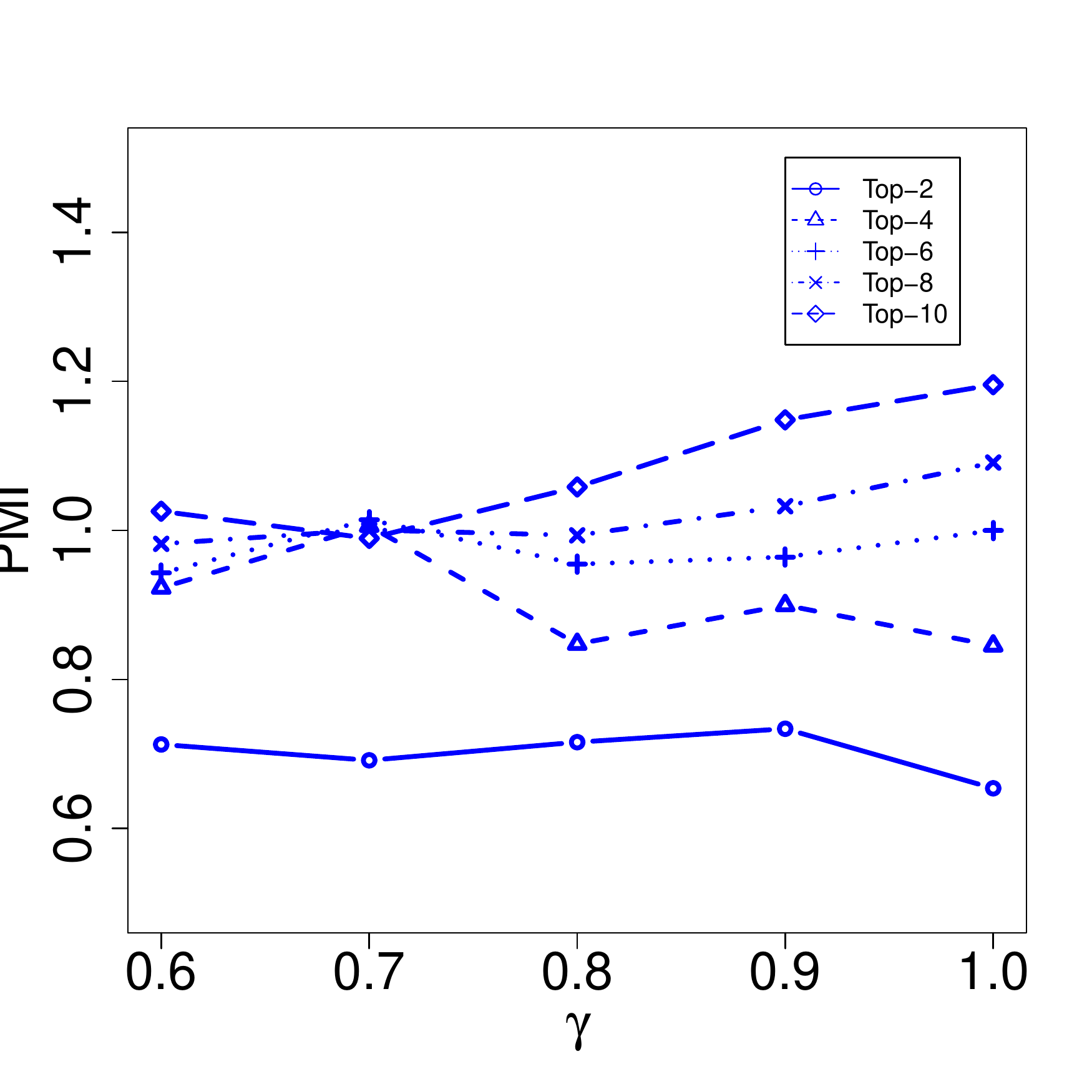}
\label{fig:subfig2}
}
\caption{Parameter sensitivity w.r.t.\ $\gamma$ on the 20NG dataset. }
\label{fig:ps-gamma-20ng}
\end{figure*}

\begin{figure*}[htdb!]
\centering
\subfigure[Perplexithy (DivPLSA)]{
\includegraphics[width=1.5in]{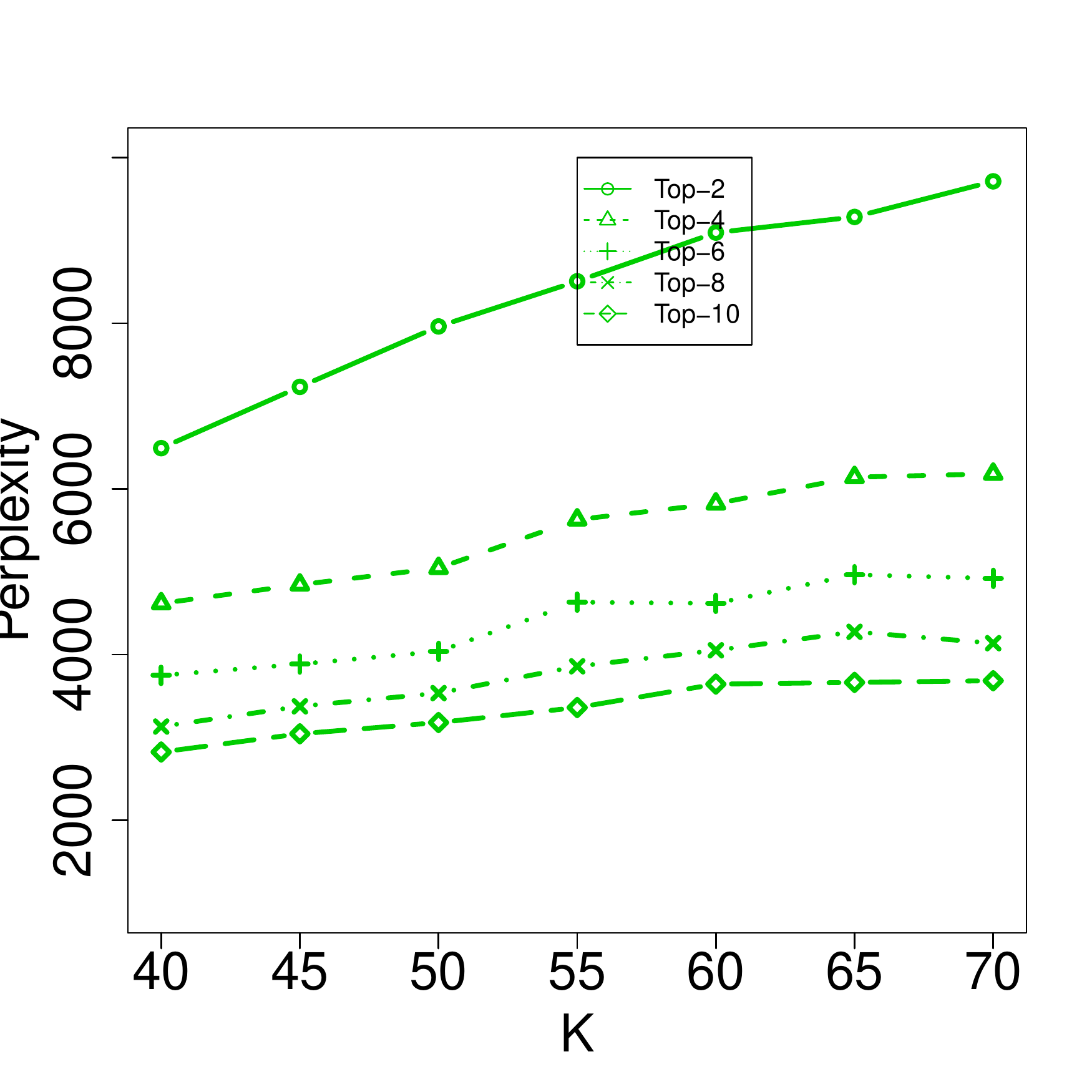}
\label{fig:subfig1}
}
\subfigure[Semantic Coherence (DivPLSA)]{
\includegraphics[width=1.5in]{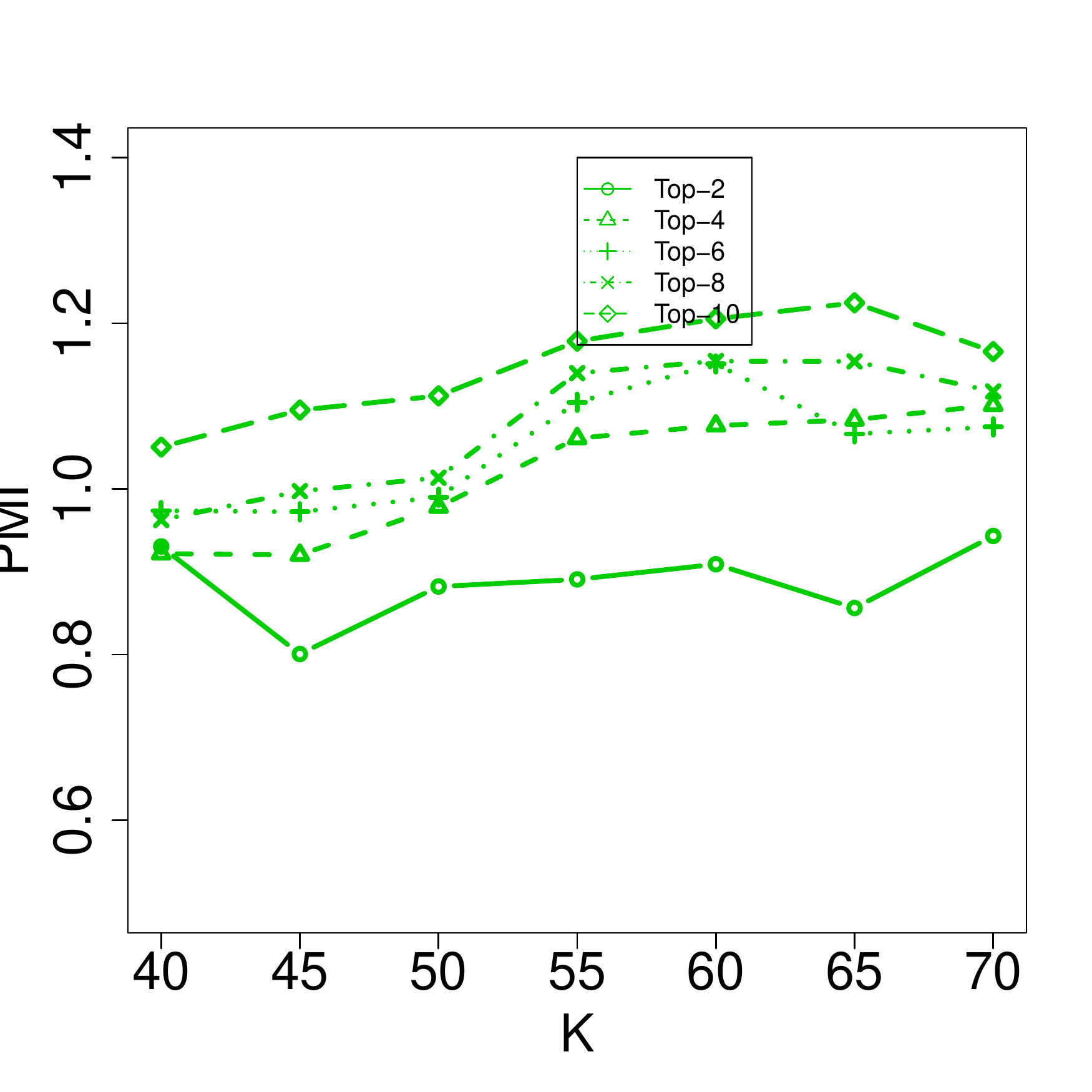}
\label{fig:subfig1}
}
\subfigure[Perplexity (DivLDA)]{
\includegraphics[width=1.5in]{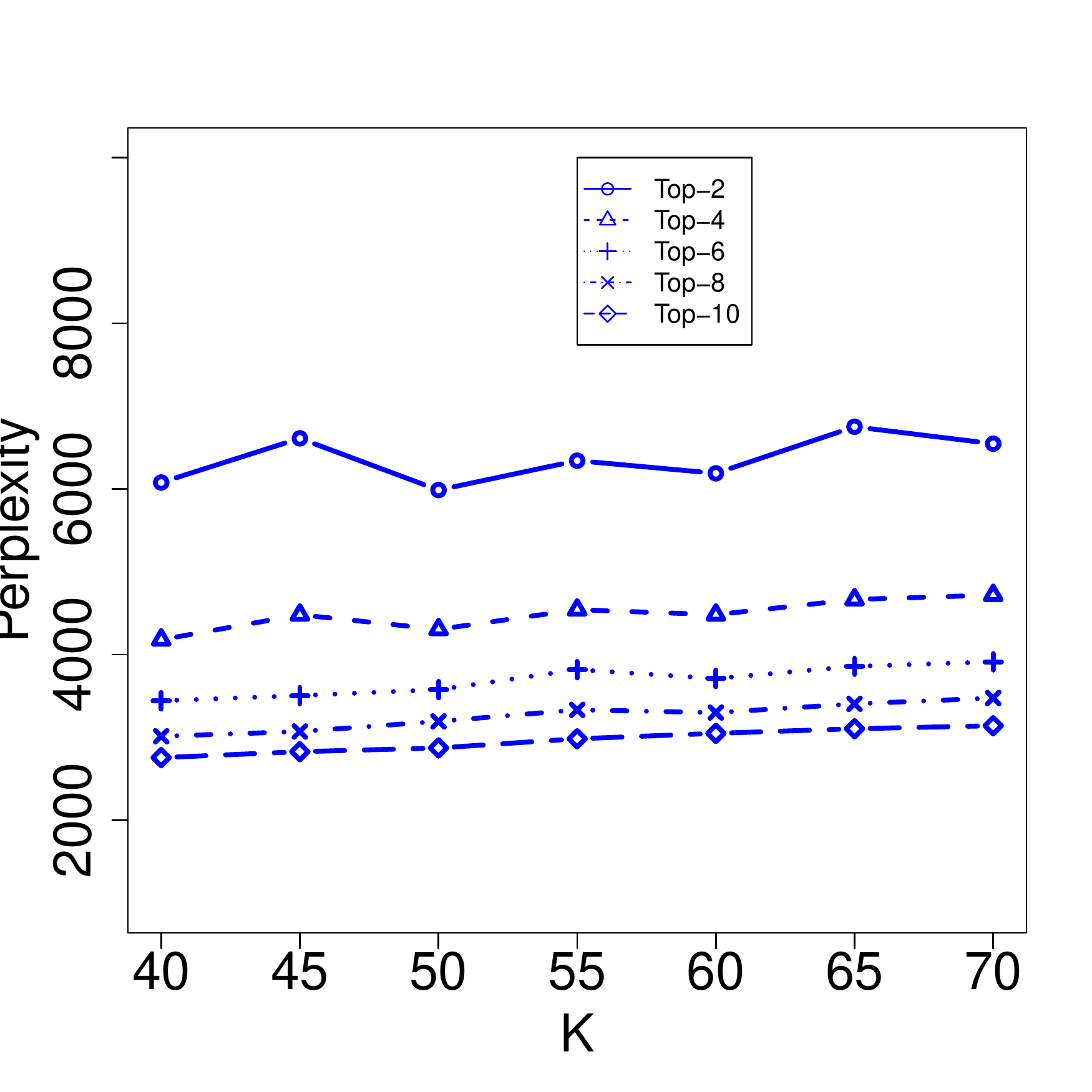}
\label{fig:subfig2}
}
\subfigure[Semantic Coherence (DivLDA)]{
\includegraphics[width=1.5in]{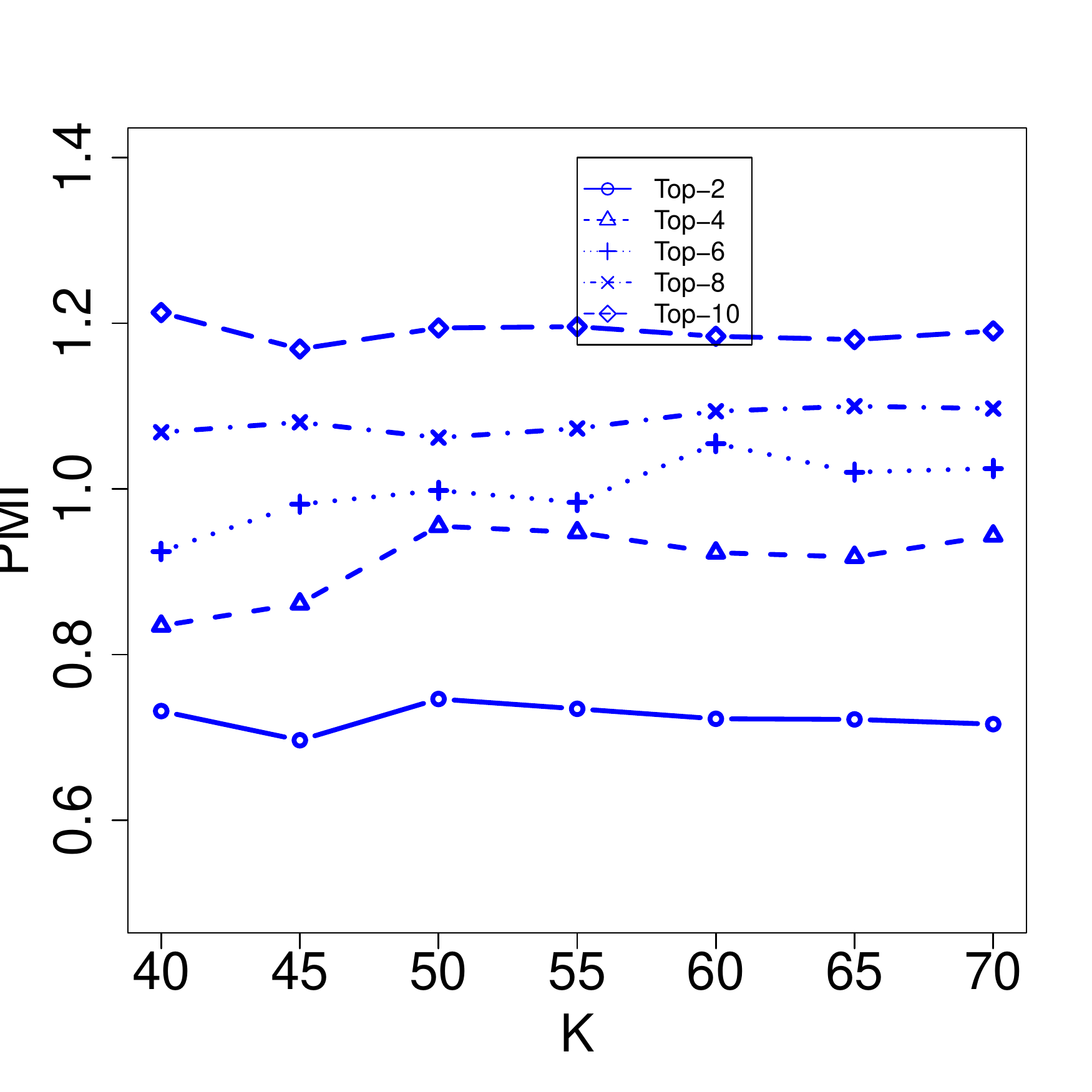}
\label{fig:subfig2}
}
\caption{Parameter sensitivity w.r.t.\ $\mathcal{K}$ on the 20NG dataset. }
\label{fig:ps-k-20ng}
\end{figure*}

In this part, we investigate the performance sensitivity of DivPLSA and DivLDA for data summarization w.r.t the parameters $\gamma$ and $\mathcal{K}$.  The parameter $\gamma$ controls the intensity of reinforcement by topic size on the transition probabilities among the topics. The larger $\gamma$ is, the more likely the larger topics will absorb the smaller topics. Therefore, the model will return fewer but larger topics with a larger $\gamma$, which tend to have a higher information coverage and lower semantic coherence. This can be observed from the results shown in Figure \ref{fig:ps-gamma-20ng}, in which the performances w.r.t $\gamma$ for both DivPLSA and DivLDA are presented.  We can see that overall the performances are not sensitive to $\gamma$. Note that DivLDA tends to use a smaller $\gamma$ than DivPLSA, this may due to the Gibbs sampling algorithm and the use of hyper-parameter $\vec{\alpha}$, which is periodically optimized.

The performance sensitivities w.r.t $\mathcal{K}$ for both DivPLSA and DivLDA in terms of  perplexity and semantic coherence  are presented in Figure~\ref{fig:ps-k-20ng}. We can see that the performances are also not sensitive to $\mathcal{K}$. 


\subsection{Summarization for DBLP}
Finally, we give a summarization for the entire DBLP dataset with our diverse topic models. The DivPLSA model is applied on the dataset with $\mathcal{K}=100$ and $\gamma=1.1$, and end up with 36 topics.  Table \ref{tab:summarization-dblp} shows the most prominent 10 topics in the dataset. 

\begin{table*}[!htdb]
\centering
\caption{Topic summarization for the DBLP dataset. }
\scalebox{0.9}{
\begin{tabular}{|c|c|c|} \hline
Human Label &Top-ranked Words& Proportion \\ \hline\hline
``methodology''&method proposed algorithm results approach based paper methods&7.69\%\\ \hline
``empirical studies''&study analysis results evaluation studies performance quality case &4.55\%\\ \hline	
``algorithm complexity''&number algorithm time complexity bound show bounds polynomial&4.17\%\\ \hline	
``knowledge representation''&model models framework knowledge modeling approach domain representation&3.98\%\\ \hline	
``computer-assisted learning''&computer learning research students project technology science social&3.81\%\\ \hline	
``machine learning''&learning classification model clustering neural training models statistical&3.65\%\\ \hline	
``parallel computing''&performance memory parallel hardware implementation applications processor processors&3.63\%\\ \hline	
``virtual systems''&system user users virtual interface interaction environment information&3.46\%\\ \hline	
\end{tabular}
}
\label{tab:summarization-dblp}
\end{table*}

\section{Conclusion and Future Work}
In this paper, we proposed two diverse topic models DivPLSA and DivLDA to learn the  prominent and diverse topics for data summarization. The two models are built on top of a reinforced random walk on the topic network, which allows the prominent topics to absorb tokens from smaller and similar topics and improves the diversity among the extracted topics. The inference procedures for the two models remain as simple and efficient as the classical ones and are appropriate for big data analysis. Experiments on four real-world datasets prove the effectiveness of the two models for data summarization.

The future work are two-fold. First, we plan to investigate the theoretical convergence of the two diverse topic models.  Currently the convergence of the two models is  empirically proved through the likelihood of the training data and the number of active topics. We believe there is an underlying objective function which tradeoffs between the data likelihood and the diversity among the topics. Second, we plan to apply the reinforced random walk into more scenarios such as unsupervised clustering, which will result in prominent and diverse clusters in the data.

\small
\bibliographystyle{abbrv}
\bibliography{sigproc}
%
%

\end{document}